%% file: arxiv-2025.tex
\documentclass[letterpaper]{article} % DO NOT CHANGE THIS
\usepackage{aaai25-arxiv}  % DO NOT CHANGE THIS
\usepackage{times}  % DO NOT CHANGE THIS
\usepackage{helvet}  % DO NOT CHANGE THIS
\usepackage{courier}  % DO NOT CHANGE THIS
\usepackage[hyphens]{url}  % DO NOT CHANGE THIS
\usepackage{graphicx} % DO NOT CHANGE THIS
\usepackage{natbib}  % DO NOT CHANGE THIS AND DO NOT ADD ANY OPTIONS TO IT
\usepackage{caption} % DO NOT CHANGE THIS AND DO NOT ADD ANY OPTIONS TO IT
\usepackage{algorithm}
\usepackage{algorithmic}

\usepackage{newfloat}
\usepackage{listings}
\DeclareCaptionStyle{ruled}{labelfont=normalfont,labelsep=colon,strut=off} % DO NOT CHANGE THIS
\floatstyle{ruled}
\newfloat{listing}{tb}{lst}{}
\floatname{listing}{Listing}
\usepackage{subfigure}
\usepackage{graphicx}
\usepackage{url}            %
\usepackage{booktabs}       %
\usepackage{amsfonts}       %
\usepackage{amsmath}
\usepackage{amssymb}
\usepackage{amsthm}         %
\usepackage{algorithm}
\usepackage{algorithmic}
\usepackage{nicefrac}       %
\usepackage{microtype}  
\usepackage{makecell}    %
\usepackage {multirow}
\usepackage[figuresright]{rotating}
\usepackage{diagbox}
\usepackage{enumitem}
\usepackage[most]{tcolorbox}

\theoremstyle{definition}

\newtheorem*{proposition*}{Proposition}

\usepackage{xcolor} % 包含xcolor包以使用颜色
\usepackage{pifont} % 包含pifont包以使用dingbat字符
\definecolor{darkgreen}{RGB}{50,100,0}
\definecolor{darkred}{RGB}{200, 0, 0}
\newcommand{\cmark}{\textcolor{darkgreen}{\ding{51}}} %
\newcommand{\xmark}{\textcolor{darkred}{\ding{55}}}
\usepackage{xcolor} % Required for defining colors
\usepackage{soul} % Required for text highlighting
\usepackage{multicol}
\usepackage[most]{tcolorbox}
\tcbuselibrary{skins}
\tcbuselibrary{breakable}
\definecolor{mintleaf}{RGB}{0, 184, 148}
\definecolor{dm-blue-500}{RGB}{0, 69, 177}
\definecolor{dm-purple-500}{RGB}{105,50,230}
\definecolor{mysilver}{RGB}{128,129,128}
\definecolor{my_green}{RGB}{0, 176, 80}
\definecolor{my_yellow}{RGB}{255,165,0}
\definecolor{my_red}{RGB}{255, 0, 0}
\definecolor{my_purple}{RGB}{126, 100, 158}
\definecolor{my_blue}{RGB}{49, 133, 155}
\definecolor{case_purple}{RGB}{160, 43, 147}
\definecolor{case_blue}{RGB}{15, 158, 213}

\title{Enhancing Decision-Making for LLM Agents via Step-Level Q-Value Models}
\author{
	Yuanzhao Zhai\textsuperscript{\rm 1,2},
	Tingkai Yang\textsuperscript{\rm 1,2},
	Kele Xu\textsuperscript{\rm 1,2},
	Dawei Feng\textsuperscript{\rm 1,2}\thanks{Corresponding author.},\\
	Cheng Yang\textsuperscript{\rm 1,2}
	Ding Bo\textsuperscript{\rm 1,2},
	Huaimin Wang\textsuperscript{\rm 1,2} \
}
\affiliations{
	\textsuperscript{\rm 1}National University of Defense Technology, Changsha, China \\
	\textsuperscript{\rm 2}State Key Laboratory of Complex \& Critical Software Environment 
}

\usepackage{bibentry}
\begin{document}

\maketitle

\begin{abstract}
Agents significantly enhance the capabilities of standalone Large Language Models (LLMs) by perceiving environments, making decisions, and executing actions. However, LLM agents still face challenges in tasks that require multiple decision-making steps. Estimating the value of actions in specific tasks is difficult when intermediate actions are neither appropriately rewarded nor penalized. In this paper, we propose leveraging a task-relevant Q-value model to guide action selection. Specifically, we first collect decision-making trajectories annotated with step-level Q values via Monte Carlo Tree Search (MCTS) and construct preference data. We then use another LLM to fit these preferences through step-level Direct Policy Optimization (DPO), which serves as the Q-value model. During inference, at each decision-making step, LLM agents select the action with the highest Q value before interacting with the environment. We apply our method to various open-source and API-based LLM agents, demonstrating that Q-value models significantly improve their performance. Notably, the performance of the agent built with Phi-3-mini-4k-instruct improved by 103\% on WebShop and 75\% on HotPotQA when enhanced with Q-value models, even surpassing GPT-4o-mini. Additionally, Q-value models offer several advantages, such as generalization to different LLM agents and seamless integration with existing prompting strategies.

\end{abstract}

\section{Introduction}
%Autonomous agents for decision-making, widely studied in reinforcement learning (RL)~\cite{sutton2018reinforcement}, have recently been proposed to be driven by large language models (LLMs) with strong reasoning and planning abilities~\cite{wang2024agentsurvey}.
%Autonomous agents driven by large language models (LLMs) are capable of performing a wide spectrum of domains across web navigation~\cite{yao2022webshop,zhou2023webarena}, interactive question answering~\cite{yang2018hotpotqa} and tool using~\cite{ma2024agentboard}.
%Using the feedback or observations from environments, LLM agents reason and plan via prompting strategies for accomplishing specific tasks~\citep{yao2022react}.
%Then the text-based outputs and action plans can be used to make API calls and perform operations within the environments.
Autonomous agents powered by large language models (LLMs) can operate across a wide range of domains, including web navigation~\cite{yao2022webshop,zhou2023webarena}, interactive question answering~\cite{yang2018hotpotqa}, and tool usage~\cite{ma2024agentboard}. By utilizing feedback or observations from environments, LLM agents can reason and plan using prompting strategies to accomplish specific tasks~\citep{yao2022react}. The resulting text-based outputs and action plans can then be employed to make API calls and execute operations within these environments.

%Recent advancements have demonstrated the potential of LLMs in various autonomous agent systems~\cite {wang2024agentsurvey}.
% , such as~\citep{deng2023mind2web}, embodied tasks~\citep{yao2022react,lin2023swiftsage} and multi-modal reasoning~\citep{Lu2023ChameleonPC}.
%Planning with feedback has been demonstrated effective in solving multi-step tasks~\citep{yao2022react}.

\begin{figure}[t]
	% \vskip 0.2in
	\begin{center}
		\centerline{\includegraphics[width=0.95\columnwidth]{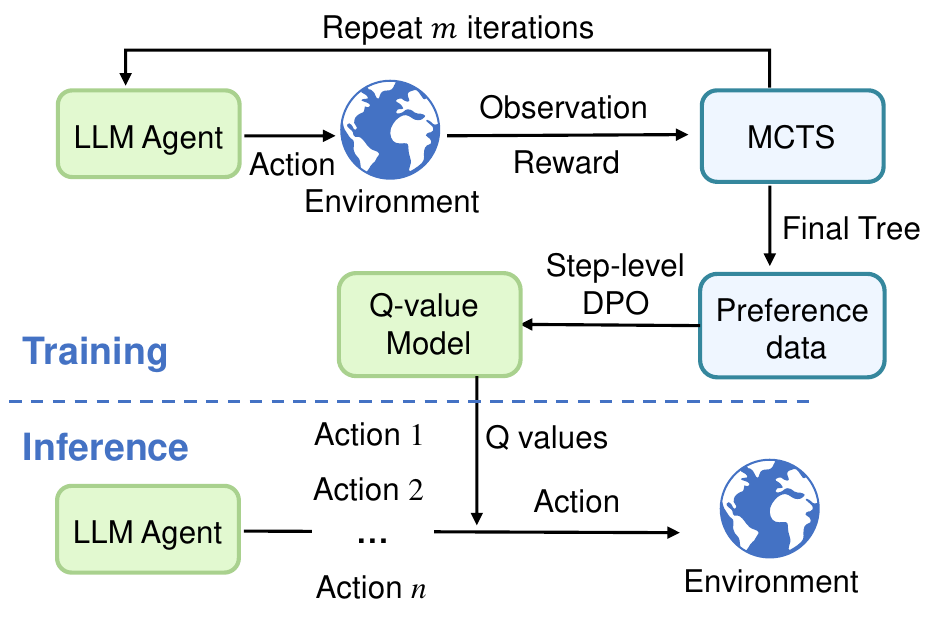}}
				\vspace{-1em}
		\caption{Overview of our method. To train the Q-value model, LLM agents interact with the environment to collect preference data with Q-value annotations using MCTS. During inference, LLM agents sample multiple candidate actions and select the best one based on the Q-value model. }
		\label{fig:Illustration}
	\end{center}
	\vspace{-2em}
\end{figure}

\begin{figure*}[!ht]
	%	\begin{center}
		%		Dyna-style~\cite{sutton1990integrated}
		\centering
		
		{
			\centering
			\includegraphics[width=0.95\linewidth]{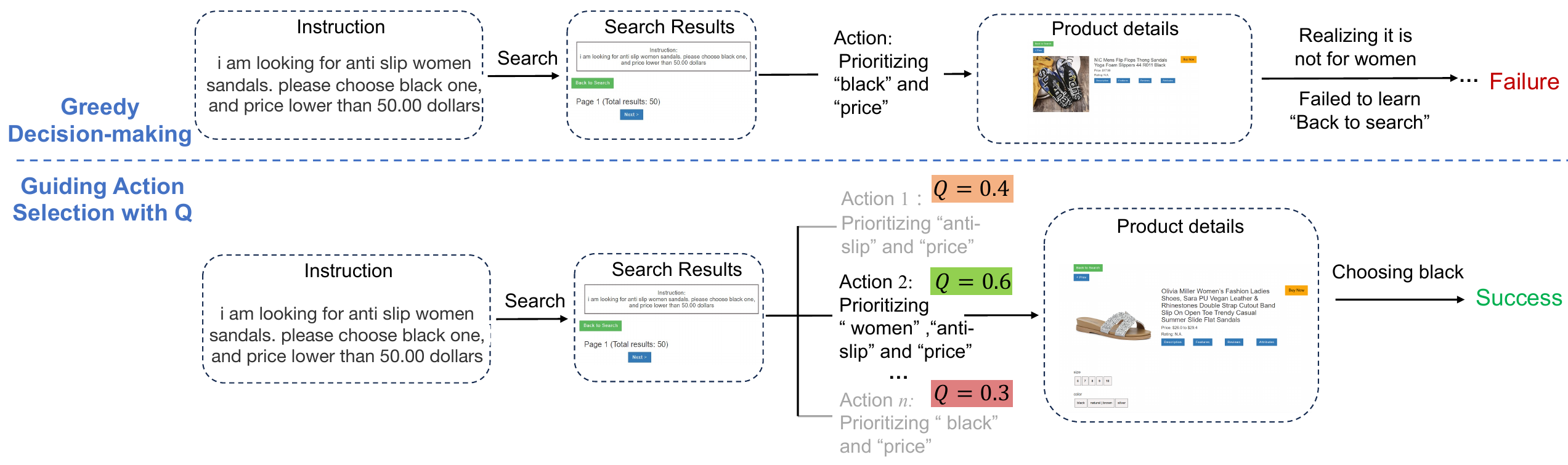}
		}
		\vskip -0.1in
		\caption{
			 Cases of GPT-4o-mini agent on WebShop. We analyze the second step of the decision-making process, where the attributes ``women," ``anti-slip," and ``price" should be prioritized over the ``black" attribute. The value of these actions is task-relevant and challenging for LLM agents to estimate. An external Q-value model can guide action selection to enhance decision-making. For further details, please refer to Appendix~\ref{appendix:webshop-case}.
%			 : actions sampled from the same LLM agent may induce different Q values for a specific task, which is task dependent and hard for LLM agents to estimate.
%			 A well-trained Q-value model can guide agents to select the most effective actions.
	\label{fig:case-study}
%			  (a) The sampled actions can be sub-optimal, which affect subsequent decision-makings and lead to failure. 
		}
		\vskip -0.1in
		%	\end{center}
\end{figure*}

Despite these advancements, even agents powered by some of the most effective LLMs, such as GPT-4, struggle with complex multi-step decision-making tasks~\citep{GPT4}. Beyond intermediate environmental feedback, additional task-specific knowledge is necessary to further enhance decision-making. Allowing LLM agents to engage in multiple trial-and-error processes during inference, strategies such as carefully designed reflection~\cite{shinn2023reflexion} or tree-based search~\cite{lats,koh2024tree} can help agents iteratively refine their actions. However, this assumption is not always feasible in realistic applications. Recently, fine-tuning open-source LLM backbones with agent trajectories has emerged as an alternative. While this approach enables LLMs to acquire more task-specific knowledge, it can also degrade their general performance~\cite{chen2024agentflan}. Furthermore, state-of-the-art API-based LLMs, which are more effective for building agents, are not accessible for fine-tuning.

%Trained with general purpose, more knowledge about specific tasks  are required to further enhance decision-making.
% to interact with the environment 
%Besides, task experience can not be accumulated with these two methods.
%To the best of our knowledge, how to guide action selection for specific environments without multiple trials and fine-tuning LLM backbones has not been studied for LLM agents.

%To enhance decision-making, recent agentic studies aim to equip LLMs trained with general purpose with more knowledge about specific tasks in addition to intermediate environmental feedback.
%In addition to intermediate environmental feedback, LLMs trained with general purpose require more knowledge about specific tasks to enhance decision-making.
%As summarized in Table~\ref{tab:method_comparison}, 
%With carefully designed reflection strategies~\cite{shinn2023reflexion} or tree-based search~\cite{lats,koh2024tree}, LLM agents

% can iterative refine their planning actions according to task completion for each instruction~\cite{huang2024understanding}.
%While effective, such approaches require multiple trails by interacting with the environment, which is prohibited in some practical scenarios.

%In many LLM agent tasks, environmental rewards are usually sparse, with only a terminal scalar indicating task completion~\cite{xi2024agentgym}.

As the number of decision-making steps increases, compounding errors and uncertainties can accumulate~\cite{xi2024training}, exacerbating the problem. Since actions are sampled from a distribution of text, the greedy action may not always be the optimal choice in the environment. As shown in Figure~\ref{fig:case-study}, suboptimal actions in intermediate steps can lead to task failure. A common and effective approach to enhancing LLMs during inference is Best-of-N sampling~\cite{yang2024asymptotics}. However, while LLM agents can sample multiple candidate actions before interacting with the environment, they often lack a clear understanding of the action values associated with task completion, as environmental rewards are typically sparse, with only a terminal scalar indicating success~\cite{xi2024agentgym}.

To overcome these limitations, we propose leveraging a Q-value model to guide action selection at each decision-making step. Q-value functions, widely adopted by traditional Reinforcement Learning (RL) agents~\cite{konda1999actorcritic,DQN}, are trained to estimate the value of specific actions. When applying the Q-value approach to LLM agents, the challenges lie in how to collect training data and how to train Q-value models effectively. As illustrated in Figure~\ref{fig:Illustration}, we integrate LLM agents with Monte Carlo Tree Search (MCTS) to iteratively explore trajectories, using its look-ahead capability to decompose sparse outcome rewards into step-level Q values. We then construct preference data based on the annotated Q-values. To train the Q-value model, we propose a step-level version of direct policy optimization (DPO)~\cite{DPO} using an additional LLM. During inference, LLM agents can sample multiple candidate actions and select the one with the highest Q value to interact with the environment in a single trial.

We conduct experiments across diverse domains, including web navigation and interactive question answering. The results demonstrate that Q-value models can clearly distinguish actions that lead to success or failure, enhancing decision-making for LLM Agents via select effective actions at each step. Additionally, task-dependent Q-value models are generalizable across different LLM agents, allowing us to utilize inexpensive LLM agents to collect training data while enhancing the decision-making of more advanced LLM agents in a plug-and-play manner. Furthermore, our method complements the design of effective prompting strategies, and integrating it with these strategies can further improve performance. In summary, our main contributions are as follows:

\begin{itemize}
%	\item We propose to utilize the MCTS algorithm to collect trajectories with Q-value annotations for LLM agents in multi-step planning tasks.
%	\item We construct step-level preference data and train a Q-value model using step-level direct policy optimization. 
	\item We leverage Q values to enhance the decision-making for LLM agents by guiding action selection at each step.
	\item We utilize the MCTS algorithm to collect decision-making trajectories and annotate them with step-level Q values.
	\item We construct preference data for training and propose step-level DPO to train Q-value models. 
	\item Experiments across two domains demonstrate the effectiveness, generalization across LLM agents, and compatibility with existing methods of our Q-value models.
%	that Q-value models can significantly enhance the planning ability of LLM agents and have several notable advantages, including generalization across different LLM agents and integration with existing prompt strategies.
\end{itemize}

%\begin{itemize}
%	\item We propose to explore high-quality trajectories via the MCTS algorithm and construct process-supervised preference data for complex multi-step tasks.
%	\item Using step-level DPO, we train a Q-value model with the preference data to distill the policy improvement of searched trajectories. 
%	\item The Q-value models can be plug-and-played into agents to improve planning performance at the inference time. Experimental results demonstrate the effectiveness and generalization of trained Q-value models for enhancing autonomous reasoning and decision-making.
%\end{itemize}
% \begin{itemize}
	%     \item We propose to utilize a step-wise verifier for the LLM agent to complete complex multi-step tasks without fine-tuning LLMs.
	%     \item We utilize both successful and failed trajectories to train verifiers iteratively, where the quality of trajectories and verifier performance are improved in each iteration.
	%     \item Our proposed iterative verifiers can be flexibly integrated with various types of API-based and open-source LLM agents.
	% \end{itemize}

\section{Related Work}
% One line of work~\citep{shinn2023reflexion,Self-refine} primarily transforms binary or scalar feedback rewards from the environment into verbal feedback, which is then incorporated as additional context into the prompt for the language agent.
% Another line of research~\citep{christianos2023pangu,song2024trial} uses trajectories and feedback to fine-tune LLM using reinforcement learning (RL) and its variants.

With the advancement of LLMs, LLM agents that interact with the world to perform a wide variety of tasks have become a major focus of research~\citep{wang2024agentsurvey}. The LLM backbone of these agents can be classified into open-source and API-based categories. Open-source LLM agents offer greater flexibility, while API-based LLMs (e.g., GPT-4) are typically more effective as agents~\cite{chen2024agentflan}. In numerous real-world scenarios, agents must execute multi-step actions to tackle complex tasks and incorporate valuable feedback to improve decision-making.

%The LLM backbones of agents can be classified into open-source and API-based.

\paragraph{Prompting Strategies.}
%Numerous prompting strategies have been proposed to improve the reasoning and planning abilities of LLM agents.
%ReAct~\citep{yao2022react} is widely used to integrate chain-of-thought (CoT)~\citep{CoT} reasoning of intermediate environment observations with agent actions.
%Reflection involves prompting an LLM to reflect on and critique the past trails to improve the current output.
%Based on reflection, self-reflexion~\citep{shinn2023reflexion} equips agents with dynamic memory and self-reflection modules. Decision-making of LLM agents can be enhanced via multiple trial-and-errors. 
%However, due to the limited input context window of LLMs, such methods are unable to accumulate extensive task experience. 

Numerous prompting strategies~\cite{wang2022self, xie2023decomposition, Self-refine} have been proposed to enhance the reasoning and planning abilities of LLM agents. In the context of enhancing decision-making, ReAct~\citep{yao2022react} is widely used to integrate chain-of-thought (CoT) \citep{CoT} reasoning with intermediate environment observations and agent actions. Reflection involves prompting an LLM to review and critique past interactions to improve current outputs. Reflexion \citep{shinn2023reflexion} provides agents with dynamic memory and self-reflection modules, enhancing decision-making through multiple trial-and-error iterations. However, due to the limited input context window of LLMs, these methods struggle to accumulate extensive task experience.

%Self-refine~\citep{Self-refine} utilizes a single LLM as a generator, refiner, and feedback provider, allowing for iterative refinement of outputs. However, it is not specifically designed for language agents.
% Retroformer~\citep{yao2023retroformer} utilizes policy gradient optimization to train a plug-in retrospective module.
% These approaches can be applied to both open-source and API-based LLMs. 
%The experiences can not be transferred even in multiple runs of the same tasks.

%\begin{table}[t]
%	\centering
%	\resizebox{\linewidth}{!}{
%		\begin{tabular}{@{}ccccc@{}}
%			\toprule
%			\multirow{1}{*}{Approach} & {\begin{tabular}[c]{@{}c@{}}Step \\ Level\end{tabular}} & {\begin{tabular}[c]{@{}c@{}}Applicable to  \\ API-based LLMs\end{tabular}}  &  {\begin{tabular}[c]{@{}c@{}}Single \\ Trial\end{tabular}} & {\begin{tabular}[c]{@{}c@{}}Task Experience \\ Accumulation \end{tabular}}  \\ \cmidrule(lr){1-5}
%			Prompting     & \xmark    & \cmark          & \cmark  or \xmark        & \xmark            \\ 
%			Fine-tuning     &\xmark  & \xmark    & \cmark      & \cmark                    \\
%			Tree Search  &\cmark  & \cmark    & \xmark       & \xmark \\
%			Q value (Ours)       & \cmark     & \cmark          & \cmark        & \cmark                  \\ \bottomrule
%		\end{tabular}
%	}
%	\caption{
%		Comparison of related work on enhancing decision-making abilities of LLM agents.
%	}
%	\label{tab:method_comparison}
%	\vspace{-0.2cm}
%\end{table}

\begin{table*}[t]
	\centering
	\resizebox{0.85\linewidth}{!}{
		\begin{tabular}{@{}p{8.8cm}cccc@{}}
			\toprule
			\multirow{1}{*}{Approach}  & {\begin{tabular}[c]{@{}c@{}}Step \\ Level\end{tabular}} & {\begin{tabular}[c]{@{}c@{}}Applicable to  \\ API-based LLMs\end{tabular}}  &  {\begin{tabular}[c]{@{}c@{}}Single \\ Trial\end{tabular}} & {\begin{tabular}[c]{@{}c@{}}Task Experience \\ Accumulation \end{tabular}}  \\ \cmidrule(lr){1-5}
			\textbf{Prompt Strategies:} Reflection, Reflexion~\cite{shinn2023reflexion} & \xmark    & \cmark          &  \cmark or \xmark        & \xmark            \\ 
			\textbf{Tree Search: } LATS~\cite{lats}, Search-agent~\cite{koh2024tree}  & \multirow{2}{*}{\cmark } & \multirow{2}{*}{\cmark}    & \multirow{2}{*}{\xmark }      & \multirow{2}{*}{\xmark} \\
			\textbf{Fine-tuning: } Agent-FLAN~\cite{chen2024agentflan}, AgentEvol~\cite{xi2024agentgym} , ETO~\cite{ETO} &\multirow{2}{*}{\xmark}  & \multirow{2}{*}{\xmark}    & \multirow{2}{*}{\cmark}      & \multirow{2}{*}{\cmark}                    \\
			\textbf{Q-value model enhanced (Ours)  }     & \cmark     & \cmark          & \cmark        & \cmark                  \\ \bottomrule
		\end{tabular}
	}
	\caption{
		Comparison of related work on enhancing decision-making abilities of LLM agents.
	}
	\label{tab:method_comparison}
	\vspace{-0.2cm}
\end{table*}

%\begin{table}[!t]  %
%	\centering
%	\small
%	\begin{tabular}{p{0.4\textwidth}P{0.11\textwidth}P{0.12\textwidth}P{0.12\textwidth}P{0.08\textwidth}}
%		\toprule
%		& Supervision-free refiner & Supervision-free feedback & Multi-aspect feedback & Iterative \\ \midrule
%		\textbf{Learned refiners}: PEER \cite{Schick2022PEERAC}, Self-critique 
%		\cite{Saunders2022SelfcritiquingMF}, CodeRL \cite{Le2022CodeRLMC}, Self-correction \cite{Welleck2022SelfCorrect}.
%		& \xmark & \cmark or \xmark  & \xmark & \cmark or \xmark
%		\\ \midrule
%		\textbf{Prompted refiners}: 
%		Augmenter \cite{Peng2023LLMAugmenter}, Re$^3$ \cite{Yang2022Re3GL}, Reflexion \cite{shinn2023reflexion}. 
%		& \cmark & \cmark or \xmark  & \xmark & \xmark 
%		\\ \midrule
%		\textbf{ours} (this work) 
%		& \cmark & \cmark & \cmark & \cmark
%		\\ \bottomrule
%	\end{tabular}
%	\caption{A comparison of ours to closely related prior refinement approaches.}
%	\label{tab:related_work_short_summary}
%\end{table}

\paragraph{Tree-based Search for LLMs.}
%Tree-based search approaches such as depth-first search (DFS), breadth-first search (BFS) and Monte Carlo Tree Search~\cite{MCTS-survey}, maintain a good exploration-exploitation trade-off in many planning algorithms~\cite{lavalle1998rapidly}.
%Equipping LLMs with tree-based search shows great potential in enhancing reasoning abilities~\cite{RAP, feng2023alphazerolike, chen2024alphamath, luo2024improve}.
%More recently, tree-based search is also integrated with LLM agents to improve planning performance.
%\citet{lats} integrate agents with MCTS, along with LM-powered value functions and other prompt-based mechanisms, such as reflection.
%\citet{koh2024tree} utilize best-first tree search to enhance LLM agents in realistic web environments.
%However, constructing a tree during inference not only introduce significant token consumption but also require the environment reversion assumption, limiting their piratical application. 

Tree-based search approaches, such as depth-first search (DFS), breadth-first search (BFS), and Monte Carlo Tree Search (MCTS)\cite{MCTS-survey}, maintain a favorable exploration-exploitation trade-off in many planning algorithms \cite{lavalle1998rapidly}. Equipping LLMs with tree-based search methods shows great potential in enhancing reasoning abilities~\cite{RAP, feng2023alphazerolike, chen2024alphamath, luo2024improve}. More recently, tree-based search has been integrated with LLM agents to improve planning performance. \citet{lats} integrate agents with MCTS, along with LLM-powered value functions and other prompt mechanisms such as reflection. \citet{koh2024tree} utilize best-first tree search to enhance LLM agents in realistic web environments. However, constructing a tree during inference not only introduces significant token consumption but also requires environmental reversion assumptions, limiting its practical application.

%Besides, they assume that the environment can be reversed in decision-making, which is not practical for some scenarios.
%However, such works significantly increase the computational overhead at inference time. 
%, show great potential in improving reasoning ability of LLMs.
%Tree-of-Thought (ToT) prompting~\cite{yao2024treeofthought} expands upon CoT by investigating multiple reasoning pathways through thoughts and employing algorithms like depth-first search (DFS) or breadth-first search (BFS) to methodically explore the tree.
%Monte Carlo Tree Search (MCTS)~\cite{MCTS-survey}, which is employed for lookahead search and instrumental in the development of the AlphaGo~\cite{alpha-go} and AlphaGo Zero~\cite{alpha-go-zero}, is recently utilized to enhance reasoning ability of LLMs~\cite{RAP, feng2023alphazerolike, chen2024alphamath, luo2024improve}.
%, both of which can find superhuman planning actions in Go.
%MCTS is demonstrated to be effective in greatly collecting high-quality process supervision data and enhancing reasoning ability~\cite{RAP, feng2023alphazerolike, chen2024alphamath, luo2024improve}.

\paragraph{Fine-tuning LLMs as Agent.}
%Besides prompting strategies, 
Fine-tuning based methods further train open-source LLM backbones as effective alternatives to API-based LLM agents.
Most fine-tuning based methods~\cite{chen2023fireact,zeng2023agenttuning,chen2024agentflan} concentrate on imitating curated expert trajectories, which is expensive and sub-optimal due to compounding errors and limited exploration data.
In order to get rid of the reliance on expert trajectories, recent works~\cite{christianos2023pangu,xi2024agentgym,ETO, zhai2024fine} collect trajectories with outcome rewards to fine-tune LLM using reject sampling fine-tuning (RFT)~\cite{yuan2023RFT}, RL or its variants.
Notably, \citet{ETO} proposes to utilize both successful and failure trajectories to fine-tune LLMs as agents via direct policy optimization (DPO)~\cite{DPO}.
%Most of these works overlook fine-grained process supervision of multi-step actions in the trajectories.
Fine-tuning LLMs with agent data on a specific tasks may deteriorate the general performance~\cite{chen2024agentflan}
Additionally, these works can not apply to API-based LLMs, which are demonstrated to be more effective in constructing agents than most open-source LLMs.

Compared to the various approaches summarized in Table~\ref{tab:method_comparison}, equipping LLM agents with step-level Q-value models offers several notable advantages. Our method can be applied to both open-source and API-based LLM agents without requiring training of the LLM backbones. Additionally, decision-making ability is enhanced by Q-values with a single trial, without needing assumptions about environmental reversion during inference. Our method does not increase context length and allows for accumulation of task experience in Q-value models, which can generalize across different agents and instructions within the task.

%Compared to different approaches summarized in Table~\ref{tab:method_comparison}, equipping LLM agents with step-level Q-value models has several notable advantages.
%Without training the LLM backbones, our method can be applied to both open-source and API-based LLM agents.
%Besides, the decision-making ability can be enhanced by Q-value with a single trial, without requiring the assumption of environment reversion during inference time.
%Our method does not incur additional context length and accumulate task experience in Q-value models, which can generalize across different agents and instructions within the task.

%We denote this advantage as single trial, which is more practical for realistic scenarios compared to tree-based and self-reflextion methods.
%is also complementary with many aforementioned approaches that develop better LLM agents.

%TODO GPT-4. RFT 
%with trajectories and outcome rewards.
%Utilizing the trajectories collected by stronger LLM agents, 

%Existing planning with feedback methods can be broadly classified into fine-tuning-free and fine-tuning-based.
%Fine-tuning-based methods~\cite{chen2023fireact,zeng2023agenttuning,chen2024agentflan,xi2024agentgym, christianos2023pangu,ETO} use trajectories with outcome feedback to fine-tune LLM using reject sampling fine-tuning, reinforcement learning (RL) or its variants.
%
%
%Besides fine-tuning-based methods, fine-tuning-free methods typically improve the agent's capabilities through prompting strategies.
% engineering or developing specialized modules. 

\section{Task Formulation}
%\section{Preliminaries}
\label{sec:Preliminaries}

% We define the collection of environments as $\mathcal{E}$. For a specific ${e} \in \mathcal{E}$, we formalize the agent task in the environment 

The agent task with environment feedback can be formalized as a partially observable Markov decision process (POMDP) $(\mathcal{U},\mathcal{S},\mathcal{A},\mathcal{O},\mathcal{T}, r)$ with instruction space $\mathcal{U}$, state space $\mathcal{S}$, action space $\mathcal{A}$, observation space $\mathcal{O}$, state transition function $\mathcal{T}: \mathcal{S} \times \mathcal{A} \rightarrow \mathcal{S}$, and reward function ${r}$.
%: \mathcal{S} \times \mathcal{A} \rightarrow \mathbb{R}$.

Given a task instruction ${u}$ in the environment, the LLM agent generates an action $a_0 \sim \pi(\cdot|u)$ based on its policy $\pi$. 
% parameterized by $\theta$
%The agent receives observation $o_1 \in \mathcal{O}$, and the state then transitions to $s_1 = [u,a_1,o_1]$.
The state then transitions to $s_1 \in \mathcal{S}$, and the agent receives observation $o_1 \in \mathcal{O}$. 
The agent continues to interact with the environment until the task is completed or the maximum number of steps is reached. 
%We adopt ReAct \citep{yao2022react} to modeling agent outputs, where the LLM agent generates reasoning thought before outputting an action.
%At time step $t$, given the history and current observation, the agent first generates the thought $h_{t+1} \sim \pi(\cdot|u,h_1,a_1,o_1,...,h_{t},a_{t},o_{t})$ first and the subsequent action $a_{t+1} \sim \pi(\cdot|u,h_1,a_1,o_1,...,h_{t},a_{t},o_{t},h_{t+1})$, which can be formulated as:
At time step $t$, given the history and current observation, the agent generates the subsequent action $a_{t+1} \sim \pi(\cdot|u,a_0,o_0,...,a_{t},o_{t})$.
%Note that when modeling outputs via ReAct \citep{yao2022react}, the action $a_{t}$ contains a reasoning thought and an executable action. 
%We adopt ReAct \citep{yao2022react} to modeling agent outputs, where the action $a_{t}$ contains a reasoning thought and an executable action.
% LLM agent generates reasoning thought before outputting an action.
Then the multi-step decision-making task can be formulated as:

%\begin{equation}
%	\begin{aligned} 
%		\pi& (\tau|u) = \prod_{t=1}^{T} \pi(a_t|s_{t-1})
%		% \\
%		% &= \prod_{t=1}^{T} \pi( a_t|u,\tau_{t-1},h_t) \cdot \pi(h_t|u,\tau_{t-1}),
%	\end{aligned}
%\end{equation}
%where we denote $T$ as the total interaction steps. 
%The initial state $s_0=[u]$, and
%$s_{t} = [u,a_{1} ,o_1,...,a_{t},o_{t}]$ represents a state comprising the original input $u$, action sequence $a_{1},...,a_{t}$, and observation sequence $o_{1},...,o_{t}$.

\begin{equation}
	\begin{aligned} 
		\pi& (\tau|u) = \prod_{t=1}^{T} \pi(a_t|u,\tau_{t-1})
		% \\
		% &= \prod_{t=1}^{T} \pi( a_t|u,\tau_{t-1},h_t) \cdot \pi(h_t|u,\tau_{t-1}),
	\end{aligned}
\end{equation}
where we denote $\tau$ as the whole trajectory, $T$ as the total interaction steps.
$\tau_{t-1}=(a_0,o_0,...,h_{t-1},a_{t-1},o_{t-1})$ denotes the interactive history up to $t-1$. 
The environment only provide the outcome reward $r(u,\tau) \in [0,1]$.
%The final reward $r(u,\tau) \in [0,1]$ is provided by the environment. 
%Rewards for immediate action is $r(s_t, a_t) = 0$ for $t < T$.
%, and $\tau_t$ as part of the interactive trajectory up to step $t$:
%\begin{align}
%	%\tau_t = (h_1,a_1,o_1,...,o_{T-1},h_{T},a_{T}) \sim \pi(\tau|e,u), 
%	\tau_t = (a_1,o_1,...,a_{t}, o_{t}) \sim \pi(\tau_t|u).
%\end{align}
%%where $e$ denotes expert interaction trajectories for few-shot examples.
%where $\tau_{t-1}=(h_1,a_1,o_1,...,h_{t-1},a_{t-1},o_{t-1})$ denotes the interactive history up to $t-1$. 
% The final reward $r(u,\tau) \in [0,1]$ is provided by the environment. 
%\subsection{Preference Learning with DPO}
%Reinforcement learning from human feedback~\cite{InstructGPT, DPO} can significantly improve the performance of LLMs for various downstream tasks.
%Recently, such methods are adopted in agent tuning to let agent learn through trial and error~\cite{ETO}.
The objective of LLM agents is to maximize rewards from the environment:
\begin{equation}
	\begin{split}
		\label{eq:objective}
		\max_{\pi}\mathbb{E}_{u\sim\mathcal{D},\tau \sim\pi\left(\cdot \vert u\right)}  \left[r(u,\tau)\right],
	\end{split}
\end{equation} 
where $\mathcal{D}$ represents the dataset containing task instructions.

%\begin{equation}
%	\begin{split}
%		\label{eq:rlhf-objective}
%		\max_{\pi}\mathbb{E}_{u\sim\mathcal{D}^u,\tau \sim\pi\left(\cdot \vert u\right)}  \left[r(u,\tau)\right]- 
%		\beta \mathbb{D}_{\mathrm{KL}}  \left[\pi (\tau\vert u)\ \vert\vert\ \pi_\mathrm{ref}(\tau\vert u)\right],
%	\end{split}
%\end{equation}
%where $\pi_\mathrm{ref}$ is the reference model, which is usually served by supervised fine-tuning LLMs. $\beta$ is the coefficient to the deviation from $\pi_\mathrm{ref}$.

\section{Proposed Method}
%In this section, we first introduce how we leverage
%Then we describe the training approach of Q-value models.
%Each planning path consists of several steps. 
% We hypothesize that not all steps in an incorrect reasoning path are equally flawed, and some may still be useful for reasoning.
%We first iteratively construct MCTS decision trees.
%To distill the values of tree nodes into the value models, we construct high-quality step-level preference data according to the decision trees and train Q-value functions leveraging step-level DPO.

%introduce an approach for  construct high-quality step-level preference data.
%Then we introduce how to distill the value 

We can build a decision tree where each node in the tree denotes an state and edge is an action.
Each node stores a set of statistics:
\begin{equation}
	\{V(s_t), N(s_t) \},
\end{equation}
where $V(s)$ represents the value function, which measures the expected reward from the sub-tree of $s_t$. $N(s_t)$ denotes the number of visits to a node $s_t$.
%The root node $s_0$ contains the full state $(e,u)$.
%Under POMDP, we can not obtain the full state but partial observation at each step $t$, thus the other nodes can be denoted by the current observations.
%We denote $r(s)$ as the reward when transfer to the state $s$, which is $0$ except for leaf nodes.

\begin{figure*}[!ht]
	%	\begin{center}
		%		Dyna-style~\cite{sutton1990integrated}
		\centering
		\subfigure[Illustration of MCTS for trajectories collection and Q-value annotation.
			\label{fig:MCTS}]
		{
			\centering
			\includegraphics[width=0.53\linewidth]{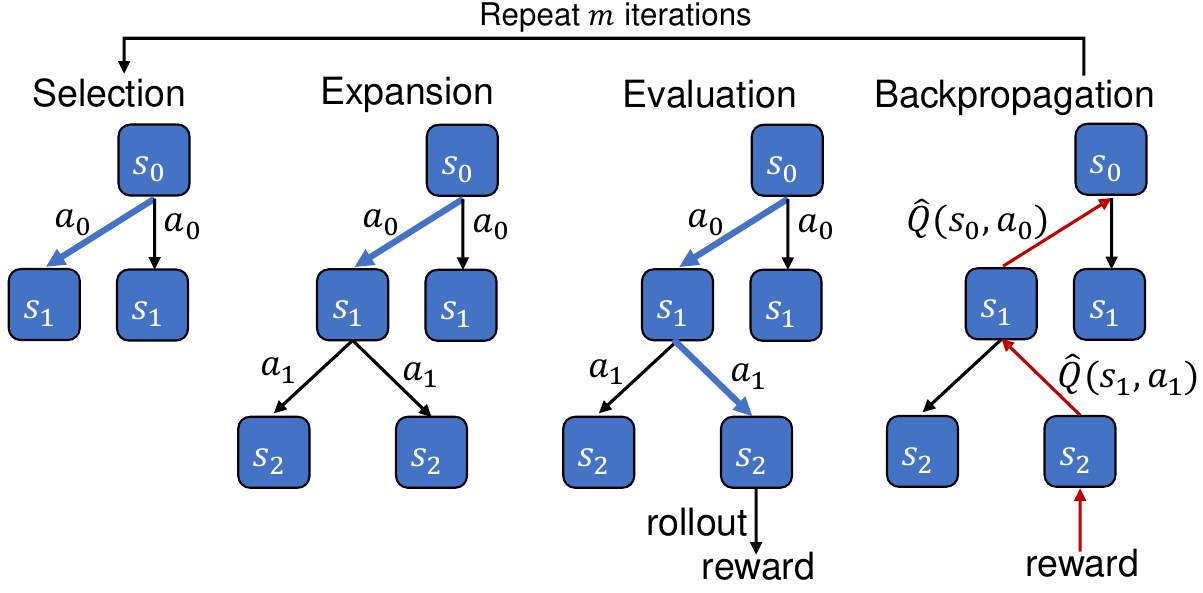}
		}
		\quad
		\subfigure[Preference data construction.
		\label{fig:preference_tree}]
		{
			\centering
			\includegraphics[width=0.35\linewidth]{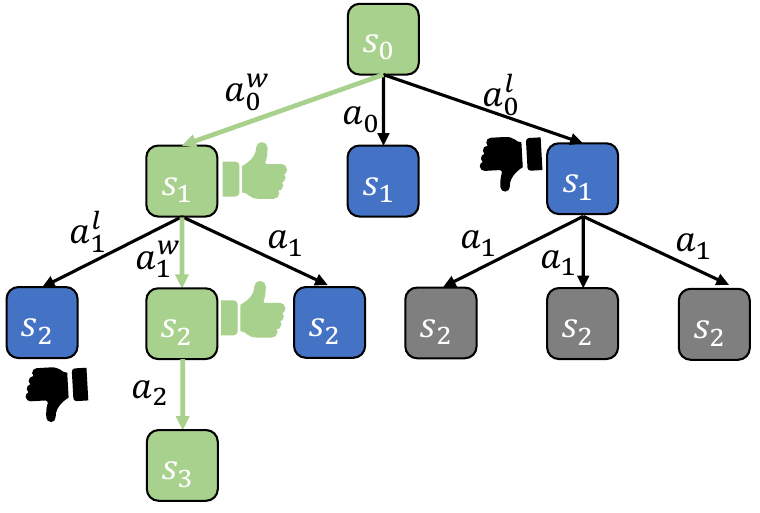}
		}
		%		\quad
		%		\subfigure[Basic idea of ORPO. 
		%		\label{fig-toy-example}]
		%		{
			%			\centering
			%			\includegraphics[width=0.32\linewidth]{figures/toy-example}
			%		}
		\vskip -0.1in
		\caption{
			Collecting step-level preference data involves two stages: (a) using MCTS to explore high-quality trajectories and annotate each step with Q-values, and (b) constructing preference data from the final tree. During the construction stage, green nodes represent the best trajectories explored by the agent and are regarded as win nodes at each depth of the tree. Blue nodes are candidates for selecting lose actions, while gray nodes are neglected.
%			Collecting step-level preference data consists of two stages: (a) MCTS to explore high-quality trajectories and annotate Q values, and (b) constructing preference data using the final tree. During constructing, reen nodes represent the best trajectory explored by the agent, which is regard as win nodes at each depth of the tree. Blue nodes are candidates for lose actions, while gray nodes are neglected. 
		}\label{fig:method}
				\vskip -0.1in
		%	\end{center}
\end{figure*}

\subsection{Step-level Q Values Estimation with MCTS}
%\subsection{Collecting Step-level Preference Data}
\label{sec:mcts}

The MCTS process starts from a root node $s_0$ and progresses through four iterative stages: selection, expansion, evaluation and backpropagation, as shown in Figure~\ref{fig:MCTS}.

\paragraph{Selection.}
The objective of the first operation, selection, is to identify the most suitable trajectories for the next expansion step.
We select the trajectory from the root node to a current leaf node.
At each depth, we select the children with the highest Upper Confidence bounds applied to Trees (UCT)~\cite{UCT} value to balance exploration and exploitation:

%用Q还是用V？ 答：Q

\begin{equation}
	\label{eq:uct}
	\small
	UCT(s_t)=V(s_t)+\sqrt{\frac{\eta \ln N\big(p(s_t)\big)}{N(s_t)}},
\end{equation}
where $\eta$ is the exploration weight, and $p(s_t)$ denotes the parent node of $s_t$.

\paragraph{Expansion.}
The second operation expands the tree by sampling $n$ actions from $\pi$, as outlined in the previous section. Unlike traditional agents, such as those used in Go, which operate in a finite action space, LLM agents have an infinite action space. LLMs can generate an unlimited number of distinct actions (sequences of tokens), though some of these may be invalid. To ensure diversity, we sample multiple candidate actions using a high temperature. The environment processes each action and provides corresponding feedback as an observation, resulting in $n$ new child nodes being added to the tree.

\paragraph{Evaluation.}
Since the tree depths for LLM agent tasks are typically much shallower than those for Go games, expansions quickly reach terminal nodes. Unlike AlphaGo~\cite{alpha-go}, which learns a value network to evaluate the value of state nodes, we evaluate the expanded nodes using a rollout algorithm. Specifically, starting from the expanded nodes, the LLM agent interacts with the environment until termination or the maximum rollout depth is reached. If the explored node is terminal, the environment's provided outcome reward is returned; otherwise, a fixed negative reward is assigned to the explored node at the maximum depth.

\paragraph{Backpropagation.}
This operation updates the tree statistics based on the outcome rewards or fixed negative rewards assigned during the evaluation stage. For each node in the trajectory $\tau$, $N(s)$ is incremented by 1, and the values are updated from the end node $s_T$ to the root node $s_0$ using the following formula:
\begin{equation}
	V(s_t) \leftarrow \frac{{V}(s_{t-1})(N(s_{t-1})-1)+r(s)}{N(s_t)}.
	\label{eq:back}
\end{equation}
The updated values are utilized in the UCT Equation~\ref{eq:uct} to guide the selection of the next node.

After multiple iterations of selection, expansion, evaluation, and backpropagation, we obtain the final tree, which stores the expanded nodes and their corresponding state values. Early stopping is triggered once the maximum reward of 1 is obtained. The Q-value of non-terminal nodes can be calculated as follows:
\begin{equation}
	\hat Q(s_t,a_t) =  r(s_t,a_t) + V(s_{t+1}) = V(s_{t+1}),
	\label{eq:Q-estimation}
\end{equation}
assuming the transition function is deterministic. Otherwise, $\hat Q(s_t, a_t)$ can be considered a Monte Carlo estimate of the true Q-value.

\subsection{Training Q-Value Models}
%Due to the limitation of MCTS iterations, $\hat Q(s_t,a_t)$ can be inaccurate to the true Q value.
%It is easier to find win and lose actions in terms of Q values among multiple candidate actions.
%Therefore, we utilize a preference leaning algorithm named direct policy optimization (DPO), and leverage its effectiveness in learning implicit value models~\cite{zhong2024dpomeetppo,DPO-q}.
%As we mentioned above, directly fine-tuning LLM backbones has several drawbacks.
%We opt for training an additional LLM $\pi_\theta$ parameterized by $\theta$ to learn Q values.
%%Considering the effectiveness of DPO in learning reward models~\cite{zhong2024dpomeetppo,DPO-q}, we train an additional LLM $\pi_\theta$ parameterized by $\theta$ to learn Q values.
%Based on the fact that evaluation tasks are simpler than generation tasks~\cite{panglanguage}, $\pi_\theta$ can be smaller than LLM backbones $\pi$ of the agent.

Due to the limitations of MCTS iterations, $\hat Q(s_t, a_t)$ may not accurately fit the true Q-value. However, it is easier to distinguish between win and lose actions based on Q-values among multiple candidate actions. Therefore, we employ a preference learning algorithm called Direct Policy Optimization (DPO), leveraging its effectiveness in learning implicit value models~\cite{zhong2024dpomeetppo,DPO-q}. As mentioned earlier, directly fine-tuning LLM backbones has several drawbacks. Instead, we train an additional LLM, $\pi_\theta$, parameterized by $\theta$, to learn Q-values. Given that evaluation tasks are simpler than generation tasks~\cite{panglanguage}, $\pi_\theta$ can be smaller than the LLM backbones $\pi$ of the agent.

Under the Bradley-Terry model~\cite{bradley1952rank}, DPO propose a preference learning loss to optimize the objective in Equation~\ref{eq:objective} while keeping the KL distance between the training model and the initial model.
\begin{equation}
	%	\resizebox{1.0\hsize}{!}{$
			\begin{aligned}\label{eq:dpo}
					&\mathcal{L}_\mathrm{trajectory}(\pi_\theta;\pi_\mathrm{ref})=
					-\mathbb{E}_{(u,\tau^w,\tau^l)\sim \mathcal{D}}\\
					&\Bigg[\log \sigma \Big(\beta \log\frac{\pi_\theta(\tau^w\vert u)}{\pi_\mathrm{ref}(\tau^w\vert u)}   
					- \beta\log\frac{\pi_\theta(\tau^l\vert u)}{\pi_\mathrm{ref}(\tau^l\vert u)}\Big)\Bigg],
				\end{aligned}
			%		$}
\end{equation} 
where $\sigma$ is the sigmoid function, $\beta$ is a weighting parameter of KL regularization,and  $\pi_\mathrm{ref}$ is the reference model, which is usually served by supervised fine-tuning LLMs before preference learning. Besides task instructions $u$. the dataset $\mathcal{D}$ contains win trajectories $\tau^w$ and lose trajectories $\tau^l$.
Without process supervision, LLM agents cannot be fine-tuned at the step level. This limitation hinders performance in multi-step decision-making tasks, as will be demonstrated in the experimental section. To address this issue, we construct more fine-grained preference data and propose a step-level version of DPO. 
%This approach makes our method more efficient and effective compared to existing fine-tuning methods.

%Directly applying DPO to fine-tune LLM backbones of agents has the following limitations.
%On the one hand, fine-tuning based methods can not be applied to API-based LLMs, which is usually more effective when acting as agents.
%On the other hand, The objective of DPO is trajectory-wise.

%To address these limitations, we train an additional LLM $\pi_\theta$ to learn Q values.

%\begin{figure}[t]
%	% \vskip 0.2in
%	\begin{center}
%		\centerline{\includegraphics[width=\columnwidth]{figures/preference_tree-v0}}
%		%		\vspace{-1em}
%		\caption{Illustration of preference data construction from the final tree. Green nodes represent the best trajectory explored by the agent, which is regard as chosen nodes at each depth of the tree. Blue nodes are candidates for constructing lose samples, while gray nodes are neglected. }
%		\label{fig:preference_tree}
%	\end{center}
%	%	\vspace{-1em}
%\end{figure}

\paragraph{Preference data construction.}

%We aim to construct step-level preference data according to $\hat Q(s_t,a_t)$ estimated by Equation~\ref{eq:Q-estimation}.
%To achieve this, we need to determine win and lose actions with the shared state.
%We first find the terminal node with the highest rewards in the final tree and then obtain the corresponding trajectories from the terminal node to the root node.
%At each depth, we take an partial segment of the selected trajectory $\tau_t$ as the shared prompt part.
%The win action $a_t^w$ are taken from the selected trajectory at the next step, while the lose $a_t^l$ action are selected from the candidate actions with the minimum Q values, as shown in Figure \ref{fig:preference_tree}.
%This approach make preference learning concentrate on distinguishing between $a_t^w$ and $a_t^l$, offering detailed insights into which action may lead to failure in the whole decision-making process, as indicated by the Q-value.

We aim to construct step-level preference data based on $\hat Q(s_t, a_t)$ estimated using Equation~\ref{eq:Q-estimation}. To achieve this, we need to identify win and lose actions for the shared decision-making trajectory segment. We first locate the terminal node with the highest reward in the final tree and then extract the corresponding trajectories from the terminal node to the root node. At each depth, we select a partial segment of the trajectory $\tau_t$ as the shared part. Win actions, $a_t^w$, are taken from the selected trajectory at the next step, while lose actions, $a_t^l$, are chosen from candidate actions with the lowest $\hat Q(s_t, a_t)$, as illustrated in Figure~\ref{fig:preference_tree}. This approach focuses preference learning on distinguishing between $a_t^w$ and $a_t^l$, providing detailed insights into which actions might lead to failure in the overall decision-making process, as indicated by the Q-value.
%The actions in the best trajectory are used for win samples.
%
%
%, which is regard as win nodes at each depth of the tree.
%This approach not only eliminates the need for labor-intensive annotation but also offers detailed insights into which action may lead to failure in the whole planning process, as indicated by the Q-value.

\paragraph{Step-level preference learning.} 
%TODO

%\begin{equation}
%	\label{eq:dpo}
%	%	\resizebox{1.0\hsize}{!}{$
%		\begin{aligned}
%			\mathcal{L}_\mathrm{DPO}(\pi;\pi_\mathrm{ref})=&
%			-\mathbb{E}_{(u,\tau^w,\tau^l)\sim \mathcal{D}}\Bigg[\log \sigma \Big(\beta \log\frac{\pi(\tau^w\vert u)}{\pi_\mathrm{ref}(\tau^w\vert u)}  \\ 
%			&- \beta\log\frac{\pi(\tau^l\vert u)}{\pi_\mathrm{ref}(\tau^l\vert u)}\Big)\Bigg],
%		\end{aligned}
%		%		$}
%\end{equation} 
%where the preference dataset $\mathcal{D}$ contains win trajectories $\tau^w$ and lose trajectories $\tau^l$.
%\citet{ETO} utilize DPO to improve LLM agents, where both win trajectories $\tau^w$ and lose trajectories $\tau^l$ are sampled from self-explored trajectories and distinguished by outcome rewards from the environment.
% with self-explored trajectories.
%success and failure trajectories during exploration.

%In this way, we avoid fine-tuning LLM backbones of agents.

%On the other hand, recent work~\cite{DPO-q} demonstrate that DPO can implicit learn credit assignment, which we also verified in the experiment section.

%Besides, in order to integrate our method with both open-source and API-based LLM agent, we train an additional LLM $\pi_\theta$ instead of fine-tuning the LLM backbones of agents.
%, we propose a step-level version of DPO to align an additional LLM $\pi_\theta$.
% train an additional LLM using step-level DPO.
Given the preference pairs $\{u, \tau_t, a_t^w, a_t^l\}$, the objective of training step-level Q-value models can be formulated as:

\begin{equation}
	\label{eq:step-dpo}
	%	\resizebox{1.0\hsize}{!}{$
		\begin{aligned}
			&	\mathcal{L}_\mathrm{step}(\pi_\theta;\pi_\mathrm{ref})=
			-\mathbb{E}_{(u, \tau_t,a_t^w,a_t^l)\sim \mathcal{D}} \\
			&\Bigg[\log \sigma \Big(\beta  \log \frac{\pi_\theta(a^w_t\vert u, \tau_t)}{\pi_\mathrm{ref}(a^w_t\vert u, \tau_t)} 
			- \beta\log\frac{\pi_\theta(a^l_t\vert u, \tau_t)}{\pi_\mathrm{ref}(a^l_t\vert u, \tau_t)}\Big)\Bigg],
		\end{aligned}
		%		$}
\end{equation} 
where $\mathcal{D}$ contains step-level preference data from $t=0$ to $t=T$.
The normalized logits of the DPO model effectively learn implicit value models~\cite{DPO, DPO-q}. In our scenario, DPO fits the estimated Q-value $\hat Q(s_t, a_t)$ and can generalize to new states and actions. With the well-trained $\pi_\theta$, the Q-value can be calculated as:
\begin{equation}
	\label{eq:Q-value}
	Q(u, \tau_t, a_t) = \beta \log \pi_\theta(a^w_t\vert u, \tau_t) - \beta \pi_\mathrm{ref}(a^l_t\vert u, \tau_t).
\end{equation} 
For brevity, we refer to $Q(u, \tau_t, a_t)$ as the Q-value model, which consists of the trained model $\pi_\theta$ and its reference model $\pi_\mathrm{ref}$ for normalization.
%Then we have the following Theorem:
%\begin{theorem}
%	\label{theorem:Q}
%	$Q(u, \tau_t, a_t)$ defined in Equation \ref{eq:Q-value} learns the optimal Q-function which models the total future reward from $s_t$ under the optimal policy.
%\end{theorem}
%这个证明用不了知乎那篇 https://zhuanlan.zhihu.com/p/693746297。那篇主要是在说用完整的句子训练后的DPO Q，会自动学到credit assignment能力；而且我们的采样策略并不是reference model，没法直接用他推。
%最后决定不证，因为我就是用的Q作为label，而不是reward
%\begin{proof}
%	Please refer to Appendix \ref{sec:proof_q}.
%\end{proof}
%During the inference time, it is not practical to assume the environmental reversion.

At inference time, the LLM agent uses the Q-value model to generate the action with the highest Q-value to interact with the environment. This is formulated as:
\begin{equation}
	\label{eq
	}
	\begin{aligned}
		a_t = \arg\max_{a}\Bigl[Q(u, \tau_t, a)\Bigl]
	\end{aligned}
\end{equation}
In practice, due to the infinite action space, we sample $n$ candidate actions, similar to the expansion stage of MCTS, and select the action with the highest Q-value to interact with the environment.

\section{Experiments}

\subsection{Experimental Settings}

To validate the versatility of our method, we apply Q-value models to various LLM backbones, including popular open-source LLMs such as the Phi-3-mini-4k-instruct model with 3.8B parameters and Llama-3.1-8B-Instruct, as well as API-based LLMs like GPT-4o-mini and GPT-4-turbo. The Q-value models are based on Phi-1.5~\footnote{huggingface.co/microsoft/phi-1\_5}, which has 1.3B parameters. For efficiency, unless otherwise stated, the LLM agents used for collecting step-level preference data are primarily based on the Phi-3-mini-4k-instruct model. The maximum context length is set to $4096$.

%
%To valid the versatile of our method, we apply Q-value models to various LLM backbones including popular open-source LLMs such as Phi-3-mini-4k-instruct model with 3.8B parameter and Llama-3.1-8B-Instruct, as well as API-based LLMs such as GPT-4o-mini and GPT-4-turbo.
%The Q-value models are established by Phi-1.5~\footnote{huggingface.co/microsoft/phi-1\_5} with 1.3B parameters.
%For efficiency, the LLM agents used for collecting step-level preference data are mainly established by  Phi-3-mini-4k-instruct model unless explicitly stated otherwise.
%The maximum context length is set to $4096$.

% because it is lightweight and support long context window.

%\paragraph{Environments.}
We evaluate our method on two tasks across different domains: WebShop~\cite{yao2022webshop} and HotPotQA~\cite{yang2018hotpotqa}. We include 3-shot in-context examples in the instruction prompt for both tasks. The maximum number of decision-making steps is set to $10$ for WebShop and $7$ for HotPotQA. For HotPotQA, we randomly select $1000$ questions for training, $100$ for validation, and $100$ for testing. For WebShop, we follow the data split described in \citet{ETO}, which consists of $1824$ instructions for training, $100$ questions for validation, and $100$ questions for testing. All experiments are conducted on a single NVIDIA A40 48G GPU, except when implementing fine-tuning-based methods, which require two NVIDIA A100 80G GPUs. Detailed information on the environment and hyperparameters can be found in Appendix~\ref{appendix:exp}.

\paragraph{Baselines.}
%We mainly compare our method with various fine-tuning based methods because both of us accumulate the task experience via training LLMs and do not require multiple trials during inference.
%Rejection Sampling Fine-Tuning (RFT)~\cite{yuan2023RFT} utilize demonstrated trajectories to train LLM backbones.
%AgentEovl~\cite{xi2024agentgym} is similar to RFT but weights trajectories according to their rewards.
%ETO~\cite{ETO} utilize DPO to improve LLM agents, where both win trajectories $\tau^w$ and lose trajectories $\tau^l$ are sampled from self-explored trajectories and distinguished by outcome rewards from the environment.
%Best-of-N (BoN) samples $n$ trajeqctories using the vanilla LLM agents and choose the one with the highest reward.
%Note that BoN serves as a strong baseline because it require query outcome reward from environments multiple times.
%The number of candidate actions are set to $n=5$ unless explicitly stated otherwise for our method and BoN.
%For fair comparision, training data for all the methods are collected from MCTS.
%ReAct-style interaction format is adopted by default, with CoT thought generated before the action.
%We also adopt another popular prompting strategy, reflection, for LLM agents to further evaluate the compatibility our Q-value models.
%\subsection{Compared to Fine-tuning-based Methods}

We mainly compare our method with various fine-tuning based methods because both approaches involve accumulating task experience through training LLMs and do not require multiple trials during inference. Rejection Sampling Fine-Tuning (RFT)~\cite{yuan2023RFT} uses demonstrated trajectories to train LLM backbones. AgentEovl is similar to RFT but assigns weights to trajectories based on their rewards. ETO employs DPO to enhance LLM agents, using both win trajectories $\tau^w$ and lose trajectories $\tau^l$, which are sampled from self-explored trajectories and distinguished by outcome rewards from the environment. Best-of-N (BoN) samples $n$ trajectories using vanilla LLM agents and selects the one with the highest reward. Note that BoN serves as a strong baseline because it requires multiple query outcome rewards from the environment. The number of candidate actions is set to $n=5$, unless otherwise specified, for both our method and BoN. For a fair comparison, training data for all methods are collected using MCTS.

\subsection{Results}
We report the results on two tasks in Table~\ref{table:results}.
As shown, our main findings are as follows:

\paragraph{Q-value models can significantly enhance decision-making.}
%The well-trained Q-value models double the performance of LLM agents established by Phi-3-mini-4k-instruct in the WebShop task and improve by 75\% in the hotpot task.
%The reason of why LLM agents can benefit more when integrated with Q-value models is two-fold.
%%It is observed that LLM agents can benefit more when integrated with Q-value models in WebShop than hotpot.
%%There are two reasons to explain the gap.
%On one hand, the WebShop task has more decision-making steps than HotPotQA, thereby using Q-value models can significantly reduce the accumulation errors.
%%tasks with more planning steps than 
%On the other hand, unlike the WebShop task that provides more distinguished rewards from $0$ to $1$, the HotPotQA provide binary rewards of $0$ or $1$. This make it harder to construct more distinguished preference data, which we will study in detailed in the next section.
%Notably, enhanced by Q-value models, Phi-3-mini-4k-instruct exceeds lightweight GPT-4o-mini on both tasks, and even exceed the most effective GPT-4-turbo on WebShop.

Well-trained Q-value models double the performance of LLM agents based on Phi-3-mini-4k-instruct on the WebShop task and improve performance by $75\%$ on the HotPotQA task. 
The enhanced LLM agent outperform the lightweight GPT-4o-mini on both tasks and even surpass the more advanced GPT-4-turbo on the WebShop task. 
There are two reasons to explain why Q-value models bring more performance gains on WebShop.
%The enhanced performance of LLM agents in WebShop compared to HotPotQA can be attributed to two factors. 
First, the WebShop task involves more decision-making steps than HotPotQA, allowing Q-value models to substantially reduce accumulation errors. Second, unlike the WebShop task, which provides more granular rewards ranging from $0$ to $1$, HotPotQA offers binary rewards of $0$ or $1$. This binary reward structure makes it more challenging to construct finely distinguished preference data, which we will explore in the next section.

\begin{table}[t]
	\centering
	\resizebox{\linewidth}{!}{
		\begin{tabular}{l|l|c|c}
			\toprule
			{\begin{tabular}[c]{@{}c@{}}\textbf{LLM} \\ \textbf{Backbone}\end{tabular}}  & {\textbf{Method}} & {\textbf{WebShop} }  &\textbf{ HotPotQA} \\
			%			&  & Success Rate $\uparrow$ &  \\
			\midrule
			%			GPT-4o-mini + BoN &  \\
			%			GPT-4o-mini + Q (ours) &  \\
			%			 \midrule[2-3]q
			\multirow{8}{*}{{\begin{tabular}[c]{@{}c@{}}Open- \\ sourced\end{tabular}}}  & Phi-3-mini-4k-instruct & 0.30  &0.20 \\
			&\ + RFT~\cite{yuan2023RFT} & 0.44 & 0.23 \\
			&\ + AgentEvol~\cite{xi2024agentgym} &0.50 & 0.23 \\
			&\ + ETO~\cite{ETO} & 0.53 & 0.27 \\
			&\ + BoN & 0.50 & 0.34 \\
			%			Phi-3 + BoN &  \\
			&\ + Q (Ours) &{0.61} (+103\%)  & {0.35} (+75\%)  \\ %T05 
			\cmidrule(lr){2-4}
			& Llama-3.1-8B-instruct & 0.48& 0.46\\
			& \ + Q (Ours) & 0.60 (+25\%) &{0.50} (+9\%) \\ %T05  
			\midrule
			\multirow{4}{*}{{\begin{tabular}[c]{@{}c@{}}API- \\ based\end{tabular}}}&GPT-4o-mini & 0.49 &  0.31 \\
			&\ + Q (Ours) & 0.64 (+31\%)  & 0.44 (+42\%)  \\
			%			\cline{2-4}
			\cmidrule(lr){2-4}
			&GPT-4-turbo & 0.58 &  0.44 \\
			&\ + Q (Ours)&  0.64 (+10\%) &   0.50 (+14\%) \\
			%			&Claude 3.5 Sonnet &  &  \\
			%			&Claude 3.5 Sonnet + Q (Ours)&  &  \\
			%			GPT-4-turbo &  \\
			%			Claude-3-5-sonnet-20240620 &  \\
			\bottomrule
		\end{tabular}
	}
	\caption{The average outcome reward of different methods on two multi-step decision-making tasks. Note that all Q-value models in this table are trained using step-level preference data collected by Phi-3-mini-4k-instruct.
		%Performance in the WebShop task are measured 
		\label{table:results}}
\end{table}

%\paragraph{Training Q-value models is more efficient and effective than fine-tuning LLM backbones.}
%RFT that utilize demonstrated trajectories to supervised fine-tune LLM improve the performances on both tasks.
%AgentEval using more reward information can further improve the performance in the WebShop task but not in the HotPotQA task.
%This is because the HotPotQA environment only provide binary rewards, reducing AgentEval to RFT.
%ETO, which adds more lose trajectory for learning, results in the best performances among fine-tuning based methods.
%These highlight the necessity of incorporating imperfect trajectories into training.
%
%Fine-tuning the LLM backbone requires high-performance computing resources especially when the LLM size and context length is large.
%Hence, our comparison with fine-tuning based methods are mainly based on Phi-3-mini-4k-instruct with 3.8B parameters.
%In comparison, our Q-value models are established by more lightweight Phi-1.5 with 1.3B parameters.
%Nevertheless, our method is more effective than all fine-tuning based methods mentioned above.
%Our method also outperforms BoN with in both tasks.
%We note that BoN is a strong baseline, which has the same computational overhead with our method but the additional outcome reward from the environment.

\paragraph{Training Q-value models is more efficient and effective than fine-tuning LLM backbones.}
RFT, which utilizes demonstrated trajectories for supervised fine-tuning of LLMs, improves performance on both tasks. AgentEval, which incorporates more reward information, enhances performance in the WebShop task but not in the HotPotQA task. This is because the HotPotQA environment only provides binary rewards, effectively reducing AgentEval's performance to that of RFT. ETO, which incorporates more losing trajectories for learning, achieves the best performance among fine-tuning-based methods. This underscores the importance of including imperfect trajectories in training.

Fine-tuning LLM backbones requires high-performance computing resources, particularly as LLM size and context length increase. Therefore, our comparison with fine-tuning-based methods primarily uses Phi-3-mini-4k-instruct with 3.8B parameters. In contrast, our Q-value models are based on the more lightweight Phi-1.5 with 1.3B parameters. Nevertheless, our method is more effective than all the fine-tuning-based methods mentioned above and outperforms BoN in both tasks.
We note that BoN, which has the same computational overhead with our method but the additional outcome reward from the environment, is a strong baseline, and our method outperforms BoN with on both tasks.

\paragraph{Q-value models are generalizable across different LLM backbones.}
The Q-value models accumulate task experience, and we expect them to generalize across different LLM agents within the same task. To verify this, we first train Q-value models using preference data sampled from Phi-3-mini-4k-instruct. We then apply these Q-value models directly to stronger open-source LLMs, such as Llama-3.1-8B-instruct, and API-based LLMs, including GPT-4o-mini and GPT-4-turbo. We observe that the decision-making abilities are consistently improved, although the performance gains are not as substantial as when the Q-value models are applied to the LLM agents that generated the training data. This is because the states and actions sampled by other LLM agents can be considered Out-Of-Distribution (OOD) relative to the step-level preference data collected by Phi-3-mini-4k-instruct, which was used to train the Q-value models.
Nevertheless, these positive results suggest that trial-and-error experience from a less powerful and more cost-effective LLM agent can benefit stronger or API-based, more expensive LLM agents.

%Results show that the average reward can be improved from $0.49$ to $0.64$.
%TODO 0.64跟用自己的Q比如何？

%\begin{table}[t]
%	\centering
%	\resizebox{0.9\linewidth}{!}{
	%		\begin{tabular}{l|c|c}	
		%			\toprule
		%			\textbf{LLM Agents} & \textbf{Average Reward $\uparrow$} & \textbf{Success Rate $\uparrow$} \\
		%			\midrule
		%			GPT-4o-mini &  0.49\\
		%			\ + Q & 0.64 \\
		%			\midrule
		%			GPT-4-turbo  &  \\
		%			\ + Q  &  \\
		%			\midrule
		%			Claude-3-5 &  \\ % -sonnet-20240620
		%			\ + Q &  \\
		%			\bottomrule
		%		\end{tabular}
	%	}
%	\caption{Applying the Q-value model trained on step-level preference data to agents with different backbones.  \label{table:generalization}}
%\end{table}

\begin{figure}[t]
	% \vskip 0.2in
	\centering
	\subfigure[Preference accuracy of Q-value models.
	\label{fig:preference_ACC}]
	{
		\centering
		\includegraphics[width=0.45\linewidth]{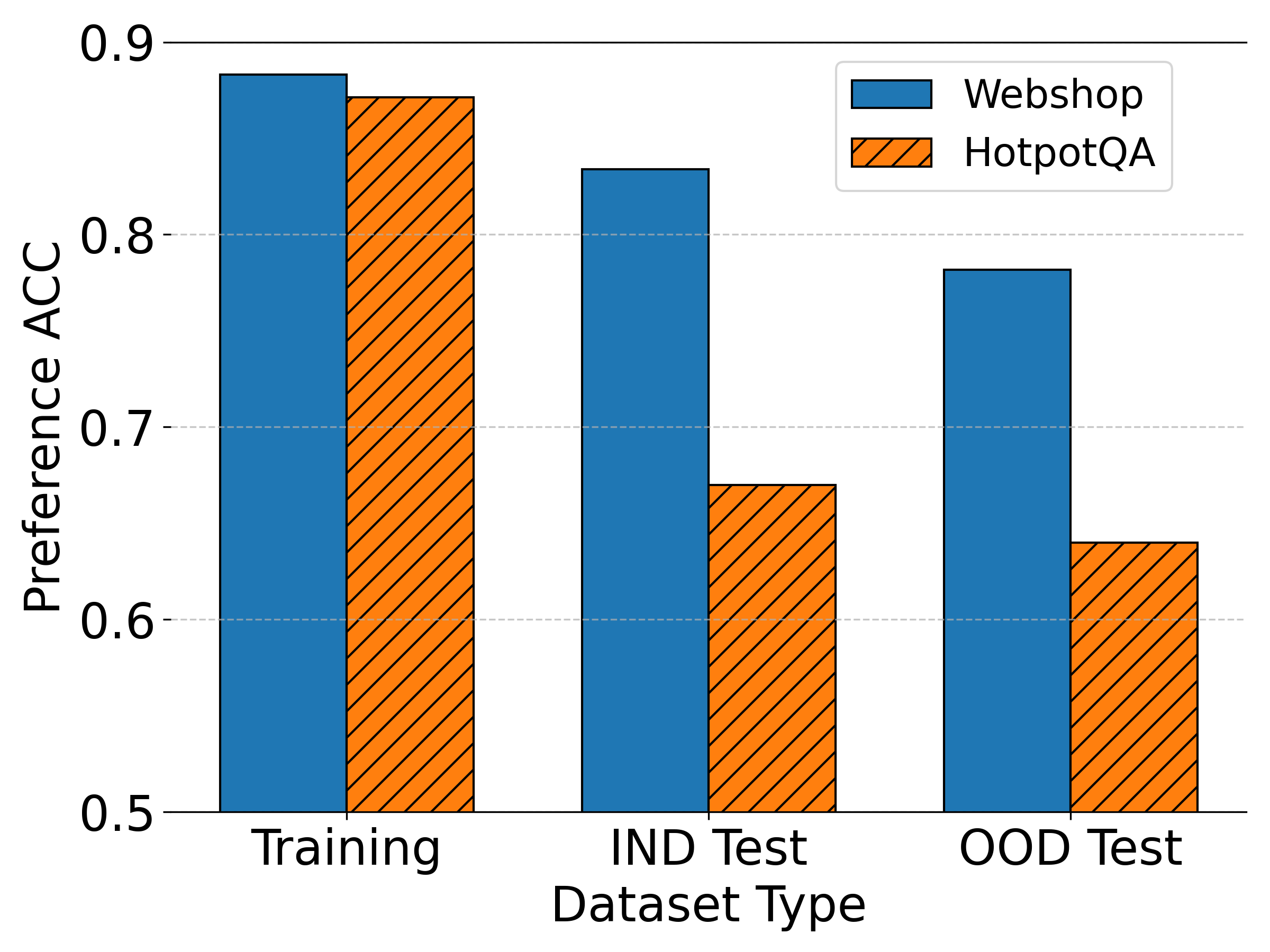}
	}
	\quad
	\subfigure[Q-value distribution of actions.
	\label{fig:Q-value-hist}]
	{
		\centering
		\includegraphics[width=0.45\linewidth]{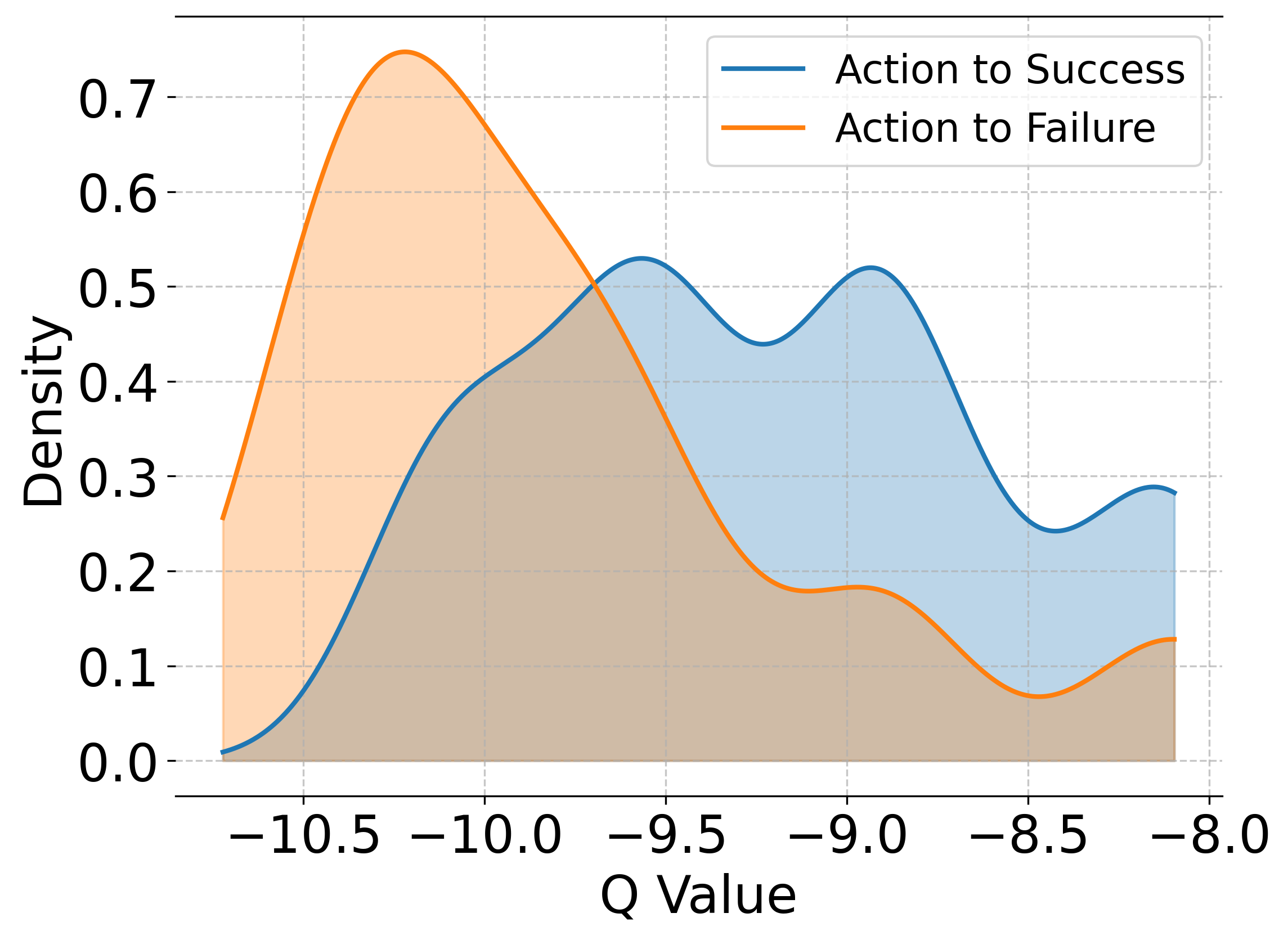}
	}
	%		\quad
	%		\subfigure[Basic idea of ORPO. 
	%		\label{fig-toy-example}]
	%		{
		%			\centering
		%			\includegraphics[width=0.32\linewidth]{figures/toy-example}
		%		}
	\caption{
		Evaluations of learned Q-value models. (a) In addition to the training and IND test datasets, we also evaluate accuracy on an OOD set, where the trajectories are sampled by the Llama-3.1-8B-instruct model. (b) We visualize the Q values of 200 actions sampled by the Phi-3-mini-4k-instruct agent, given the instructions in the test set of WebShop.
	}
	%	\vspace{-1em}
\end{figure}

%\subsection{Evaluations of Q-value Models}
%We further investigate the accuracy of Q-value models in assessing preference relationships of collected step-level data.
%As shown in Figure~\ref{fig:preference_ACC}, it is observed that the preference relationships within the training sets can be easily learned in both tasks.
%When evaluating on in-distribution (IND) test set, the accuracy decrease to $83\%$ on WebShop and $67\%$ on HotPotQA.
%The performance gap on HotPotQA is due to its binary outcome reward and the early stopping of MCTS when obtaining the reward $1$.
%Besides, generalizing to OOD test set in which preference data is collected by other LLM agents causes a slight performance degradation on both tasks.
%Nevertheless, such preference accuracy is enough to enhance the performance of the downstream tasks, which is consistently with recent studies of learning reward models~\cite{lambert2024rewardbench}.

%说一下为啥hotpot不够牛逼
\subsection{Evaluations of Q-value Models}

We further investigate the accuracy of Q-value models in assessing the preference relationships of collected step-level data. As shown in Figure~\ref{fig:preference_ACC}, preference relationships within the training sets are learned effectively in both tasks. However, when evaluating on the in-distribution (IND) test set, accuracy decreases to $83\%$ on WebShop and $67\%$ on HotPotQA. The performance gap on HotPotQA is attributed to its binary outcome reward and the early stopping of MCTS when the reward of $1$ is obtained. Additionally, generalizing to the OOD test set, where preference data is collected by other LLM agents, results in a slight performance degradation on both tasks. Nevertheless, this level of preference accuracy is sufficient to enhance the performance of downstream tasks, consistent with recent studies on learning reward models~\cite{lambert2024rewardbench}.

%To further evaluate the effectiveness of Q-value Models, we select 200 actions from the successful and failed trajectories respectively, and visualize their Q values in Figure~\ref{fig:Q-value-hist}.
%The Q-value distribution for actions in failed trajectories is skewed towards left.
%In contrast, the distribution for incorrect solutions shows less skewness, with most of the probability density leaning towards right. 
%This pattern may arise because failure often results from choosing some destructive actions~\cite{koh2024tree}, which suggests that our Q-value models have the capability of credit assignment.
To further evaluate the effectiveness of Q-value models, we select 200 actions from successful and failed trajectories, respectively, and visualize their Q-values in Figure~\ref{fig:Q-value-hist}. The Q-value distribution for actions in failed trajectories is skewed to the left, while the distribution for successful actions shows less skewness, with most of the probability density leaning to the right. This pattern may arise because failures often result from choosing detrimental actions~\cite{koh2024tree}, suggesting that our Q-value models are capable of effective credit assignment.

\subsection{Ablation Studies}

\begin{table}[t]
	\centering
	\resizebox{\linewidth}{!}{
		\begin{tabular}{l|c|c|c|c}
			\toprule
			\textbf{Preference Data} & $n=1$ & $n=3$ & $n=5$& $n=7$ \\
			\midrule
			Step-level & 0.30& 0.50 &0.61 & 0.63 \\
			Trajectory-level & 0.30& 0.42 & 0.50 & 0.51 \\
			\bottomrule
		\end{tabular}
	}
	\caption{Average rewards of LLM agents powered by Phi-3-mini-4k-instruct on WebShop. \label{table:step-trajectory}}
\end{table}

%\subsection{Advantage of Step-level Preference Data}
%\paragraph{Advantage of step-level preference data.}
%Recent works~\cite{DPO-q, zhong2024dpomeetppo} show that trajectory-level DPO objective in Equation~\ref{eq:dpo} also has the potential in credit assignment.
%Therefore, we set another baseline to compare our proposed step-level Q-value model with the Q-value model trained with trajectory-level preference data $(u,\tau^w,\tau^l)$.
%Our results in Table \ref{table:step-trajectory} suggest that Q-value models trained with trajectory-level data can also enhance LLM agents, whose performance gradually improves as more candidate actions in each step are sampled.
%However, training with our step-level preference data outperforms the baseline in various sampling numbers of candidate actions.
%This can be attributed to the more granular information of planning steps represented by the node value in the Monte Carlo tree.
%Specifically, we first sample multiple complete trajectories.
%According to the outcome reward, we select the trajectory with highest reward as win, while select the trajectory with lowest reward as lose.

\paragraph{Advantage of Step-Level Preference Data.}

Recent studies~\cite{DPO-q, zhong2024dpomeetppo} indicate that the trajectory-level DPO objective, as described in Equation~\ref{eq:dpo}, also holds potential for credit assignment. To evaluate this, we establish an additional baseline by comparing our proposed step-level Q-value model with a Q-value model trained using trajectory-level preference data $(u, \tau^w, \tau^l)$. Our results, as shown in Table~\ref{table:step-trajectory}, suggest that while Q-value models trained with trajectory-level data can enhance LLM agents, their performance improves gradually as more candidate actions are sampled at each step. However, models trained with our step-level preference data consistently outperform this baseline across various numbers of candidate actions. This superior performance can be attributed to the more granular information provided by planning steps, as represented by the node values in the Monte Carlo tree.

\begin{figure}[t]
	% \vskip 0.2in
	\centering
	\subfigure[Number of training preference data on performance.
	\label{fig:training_num}]
	{
		\centering
		\includegraphics[width=0.45\linewidth]{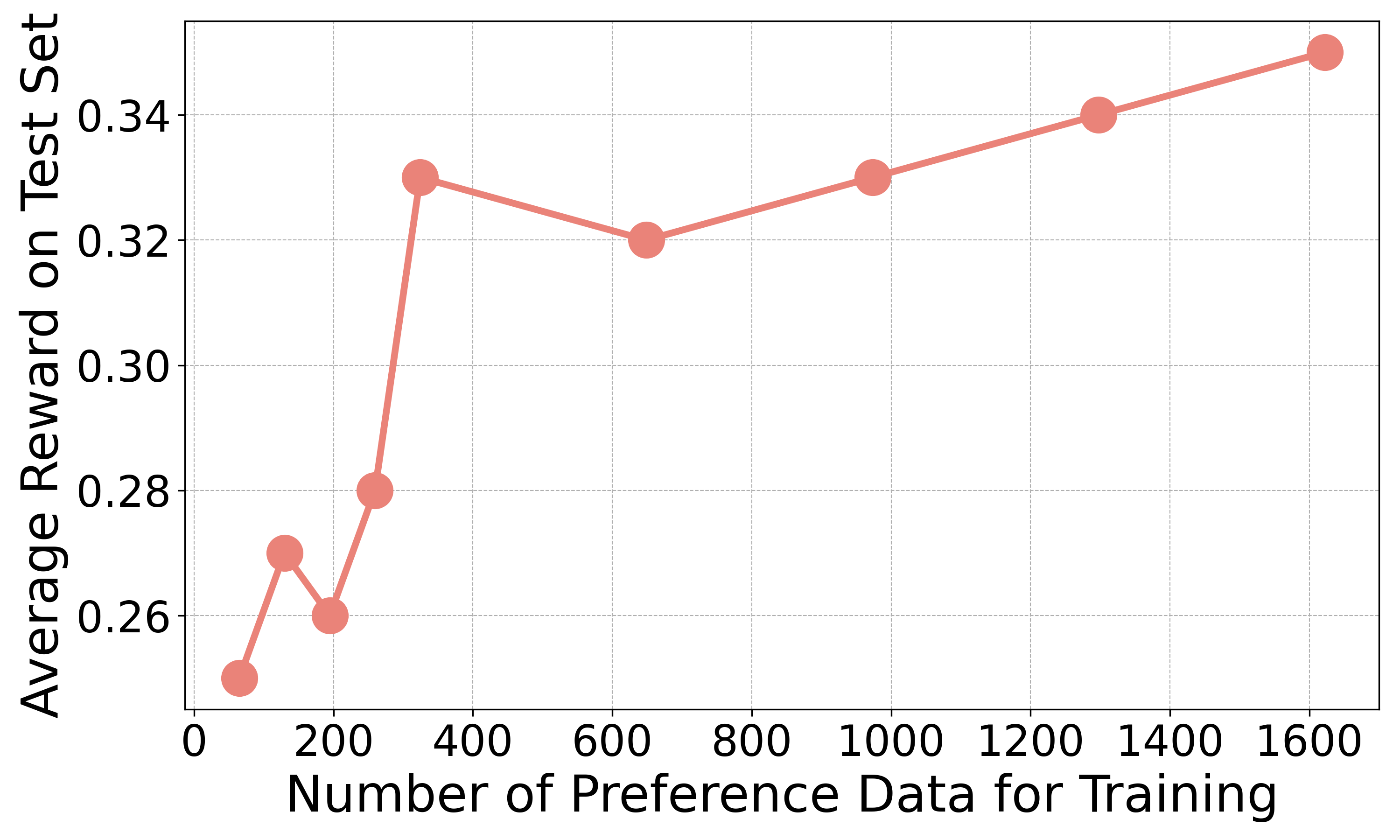}
	}
	\quad
	\subfigure[Preference data construction with different MCTS iterations.
	\label{fig:MCTS_iterations}]
	{
		\centering
		\includegraphics[width=0.45\linewidth]{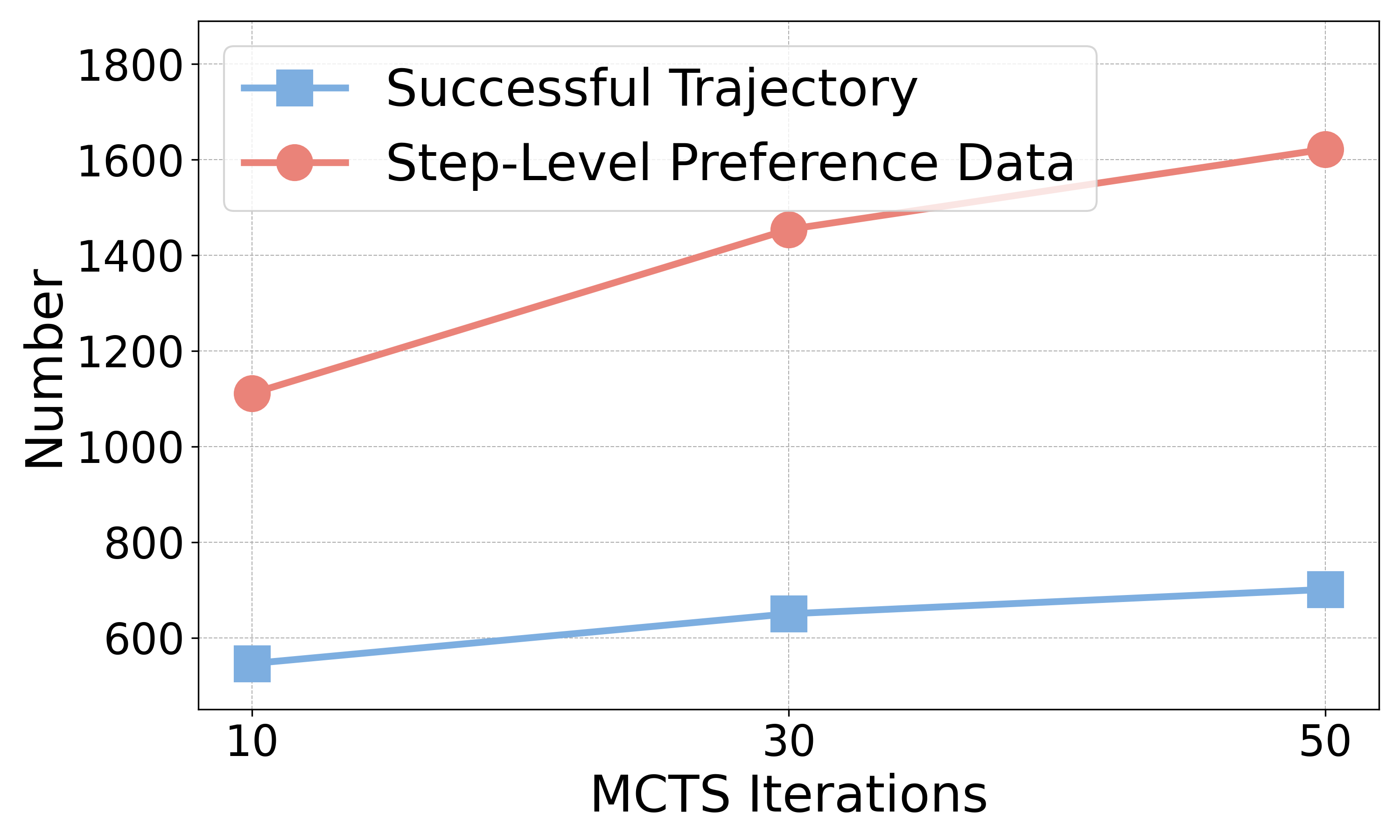}
	}
	%		\quad
	%		\subfigure[Basic idea of ORPO. 
	%		\label{fig-toy-example}]
	%		{
		%			\centering
		%			\includegraphics[width=0.32\linewidth]{figures/toy-example}
		%		}
			\vspace{-1em}
	\caption{
		Ablations of training samples and collection of preference data.
	}
		\vspace{-1em}
\end{figure}

%\paragraph{How many preference data do we need for training?}
%To train a Q-value model, we need to utilize the task instructions to construct step-level preference data.
%We explore the impact of different training sizes on their downstream performance.
%As shown in Figure~\ref{fig:training_num}, we evaluate several checkpoints of one epoch when training the Q-value model on the HotPotQA task, which can represent different training sample usages.
%We observe that less than $400$ step-level preference data can significantly boost the performance, which can be collected with only around $250$ task instructions in our setting.

\paragraph{How much preference data is needed for training?}
To train a Q-value model, step-level preference data must be constructed using task instructions. We investigate how different amounts of training data impact downstream performance. As shown in Figure~\ref{fig:training_num}, we evaluate several checkpoints from one epoch of training the Q-value model on the HotPotQA task, which represents varying quantities of training samples. 
We observe that fewer than $400$ step-level preference data points can significantly enhance performance, achievable with approximately $250$ task instructions in our setting.
This demonstrates the sample efficiency of our approach for training Q-value models.

%\paragraph{Ablation of MCTS iterations.}
%One can also collect more preference data through more MCTS iterations, at the cost of increasing computational overhead.
%In our previous experinemts, we set the MCTS iteration $m=30$ by default.
%We ablate the iteration of MCTS to show its effect on data collection.
%As shown in Figure~\ref{fig:MCTS_iterations}, the number of successful trajectories that can be used to construct step-level preference data increases as we enlarge the maximum MCTS iteration.
%It is observed that nearly all MCTS processes stop early before the 50th iterations due to obtaining the maximum reward. Therefore, it is meaningless to continue increasing the number of iterations.
%Besides, the number of step-level preference data increase faster than success trajectory as MCTS iterations. This is because trajectories explored with large MCTS iterations usually have more decision-making steps to construct step-level preference data. 

%It is worth to note that some MCTS process stop early due to obtaining the maximum reward.
%We observe that due to our early stopping mechanism, many MCTS process can 

\paragraph{Ablation of MCTS Iterations.}
More preference data can be collected by increasing the number of MCTS iterations, though this also increases computational overhead. In our previous experiments, we set the MCTS iteration to $m=30$ by default. We perform an ablation study on the number of MCTS iterations to assess its impact on data collection. As shown in Figure~\ref{fig:MCTS_iterations}, the number of successful trajectories available for constructing step-level preference data increases with the maximum number of MCTS iterations. Nearly all MCTS processes terminate early, before the 50th iteration, due to achieving the maximum reward or depth, rendering additional iterations redundant. Furthermore, the number of step-level preference data points increases more rapidly than the number of successful trajectories with additional MCTS iterations. This is because trajectories explored with a larger number of MCTS iterations typically involve more decision-making steps, thus providing more step-level preference data.

\begin{table}[t]
	\centering
	\resizebox{0.85\linewidth}{!}{
		\begin{tabular}{l|c}
			\toprule
			\textbf{Method} & \textbf{HotPotQA}  \\
			\midrule
			ReAct  & 0.31 \\
			ReAct + Reflection &  0.39\\
			%			Reflexion {\scriptsize~\citep{shinn2023reflexion}} &  \\
			%			ReAct (Best of N)  & \\
			%			Self-Reflection With Q (Ours) \\
			ReAct + Q (Ours) & 0.46 \\
			ReAct + Reflection + Q (Ours) & 0.48 \\
			%			ToT {\scriptsize~\citep{yao2024treeofthought}} &  \\
			%		RAP &  \\ 
			%			RAP {\scriptsize~\citep{RAP}} & \\
			%			LATS {\scriptsize~\citep{lats}} & \\
			%			With Q + Reflexion (Ours) &  \\
			\bottomrule
		\end{tabular}
	}
	\caption{Averaged rewards of integration with different prompting strategies. \label{table:reflection}}
		\vspace{-1.5em}
%	\caption{Ablations of various prompting strategies integrated with our method.\label{table:reflection}}
\end{table}

\paragraph{Integration with different prompting strategies.}
%In our work, we adopt the ReAct-style prompt to make LLMs to act as agents.
%We further equip LLM agents with more sophisticated prompting strategy, ``ReAct + Reflection''.
%As shown in Table~\ref{table:reflection}, the performance of GPT-4o-mini can be improved from $0.31$ to $0.39$.
%We tried to equip the prompting strategy with the LLM agent established by Phi-3-mini-4k-instruct.
%Unfortunately, the performance degradation from $0.2$ to $0.15$.
%This may be because Phi-3-mini-4k-instruct with only 3.8B parameters can not well understand the reflection prompts.
In our work, we use a ReAct-style prompt to enable LLMs to function as agents. We further enhance LLM agents with a more sophisticated prompting strategy, ``ReAct + Reflection''. As shown in Table~\ref{table:reflection}, this improves the performance of GPT-4o-mini from $0.31$ to $0.39$. We also apply the prompting strategy to the LLM agent based on Phi-3-mini-4k-instruct. However, the performance decreased from $0.20$ to $0.15$. This may because that Phi-3-mini-4k-instruct with 3.8B parameters can not adequately understand the reflection prompts.

%For API-based LLM agents, the most effective approach to improve their decision performance could be utilizing more effective prompting strategies.
%We first verify this with the commercial API-based LLM agent, GPT-4o-mini.
%To verify the compatibility abilities of our Q-value models, we integrate our method with the two prompting strategies.
%We use the same experimental settings as Table~\ref{table:results} to train Q-value models, except that we adopt various prompting strategies and sample trajectories using GPT-4o-mini instead of Phi-3-mini-4k-instruct.
%Results show than methods with both reflection and Q-value models achieve the best average reward of 0.48, suggesting that our proposed method is complementary with designing more effective prompting strategies.
%Besides, combined the results in Table \ref{table:results} and \ref{table:reflection}, we observe that the Q-value model trained on preference data collected by itself is more effective than one trained on data sampled by Phi-3-mini-4k-instruct, with which the average reward is $0.46$ and $0.48$ respectively.
%This is consistently with our observations in terms of preference accuracy on IND and OOD test set in Figure~\ref{fig:preference_ACC}.

We use the same experimental settings as described in Table~\ref{table:results} to train Q-value models, but with different prompting strategies and by sampling trajectories using GPT-4o-mini instead of Phi-3-mini-4k-instruct. The results indicate that methods incorporating both reflection and Q-value models achieve the highest average reward of $0.48$, suggesting that our proposed method complements the design of more effective prompting strategies. Additionally, combining the results from Table~\ref{table:results} and Table~\ref{table:reflection}, we observe that the Q-value model trained on preference data collected by GPT-4o-mini outperforms the model trained on data sampled by Phi-3-mini-4k-instruct, with average rewards of $0.48$ and $0.46$, respectively. This finding is consistent with our observation that the preference accuracy on the OOD test set exceeds the preference accuracy on the IND test set, as shown in Figure~\ref{fig:preference_ACC}.

%\subsection{Evaluating Q-value Models as a Generator}
%Since our Q-value models are also LLMs, we also evaluate its agent abilities for generator. Unfortunately, 

%\subsection{Token consumption During Inference}

%\subsection{Ablation Studies}

%\paragraph{Ablation of Training Sample Numbers}
%1000 - +
%(Webshop)
%
%\paragraph{Increasing Training Samples or MCTS iterations}
%MCTS iterations
%
%
%\section{Discussions, Limitations and Future Work}
%For fair comparison, we do not compare our proposed methods with multi-trial methods such as reflexion and LATS~\cite{lats} which require querying the environment several times for outcome reward.
%Compared to such methods, our method does not require the assumption of environment reversion during inference time.
%Though we still assume a reversed environment after the evaluation stage of the MCTS process for collecting preference data, we believe it is more practical because we can collect data in simulated environments and generalize trained Q-value models for practical use.
%Our method is also complementary with these methods. 

%1B 的模型

%We utilize the naive MCTS to collect step-level preference data.

\section{Conclusion and Limitations}
In this paper, we propose leveraging Q-values to guide action selection at each decision-making step. We collect training data using MCTS and train Q-value models through step-level direct policy optimization. Results from two distinct tasks demonstrate that our method is more efficient and effective compared to fine-tuning LLM backbones. Furthermore, the trained Q-value models are plug-and-play, easily applicable to both open-source and API-based LLM agents, and generalize well across them. We believe our method introduces a novel and flexible paradigm for enhancing the decision-making capabilities of LLM agents.
%, complementing existing agentic methods.

While collecting training data introduces $\mathcal{O}(kn)$ sample complexity, the feasibility of sampling with lightweight open-source LLM agents makes this manageable. Our method does not increase context length, but it does introduce $n$-fold token consumption for sampling multiple candidate actions during inference. This trade-off is acceptable and can be further optimized through caching technologies. Due to computational resource constraints, the Q-value models are limited to 1.3B parameters. Exploring the use of more powerful LLMs could enhance the effectiveness of Q-value models, which we plan to address in future work.

%Multi-round preference data collection via MCTS

% Entries for the entire Anthology, followed by custom entries
\bibliography{custom}
%\bibliography{aaai25}

\newpage
\clearpage
\newpage
\include{appendix}

\end{document}

%% file: appendix.tex
\appendix
\setcounter{secnumdepth}{2}
\onecolumn

%\section{Proof of Theorem~\ref{theorem:Q}}
%\label{sec:proof_q}

\section{Experimental Setup Details}
\label{appendix:exp}
\subsection{Environment Details}
\paragraph{WebShop.}
WebShop tasks the agent with solving a shopping task by browsing websites with detailed product descriptions and specifications. The available action APIs include search[QUERY] for using the search bar and click[BUTTON] for clicking buttons on web pages. Clickable buttons include product titles, options, buy, back to search, and previous/next page, among others. When the agent selects the ``Buy Now" action, the environment provides \textbf{an outcome reward ranging from $0$ to $1$} based on the matching heuristics of the product's attributes and price.

\paragraph{HotPotQA.}
HotPotQA is a question-answering task that requires retrieval across Wikipedia passages. Following the setup of \cite{yao2022react}, LLM agents are equipped with API calls for searching (Search[INFORMATION]) and retrieving (Lookup[INFORMATION]). Upon receiving an answer, the environment provides \textbf{a binary outcome reward of $0$ or $1$} based on its correctness according to the ground truth.

\subsection{Hyper-parameters}
\label{appendix:hyper-parameter}
%We set the maximum iterations of MCTS to $m=30$ by default to collecting step-level preference data, with $n=5$ expansion width and the exploration weight $\eta$ is set to 2.
%The sampling temperature of LLM agents is set to $1.0$ to encourage diversity.
%
%The batch size is $16$, and the learning rate is set to $1e-5$ with $0.1$ warm-up ratio and a cosine scheduler.
%The $\beta$ in DPO loss is set to $0.1$.
%We train Q-value models up to $3$ epochs and choose the checkpoints that have the best performance on the validation set, which induces $1$ epoch on HotPotQA and $2$ epoch on WebShop.

%For training Q-value models, we use Phi-1.5~\footnote{huggingface.co/microsoft/phi-1\_5} with 1.3B parameters because it is lightweight and support long context window.
%The hyper-parameters of training Q-value models can be found in \cref{label}.

%The maximum epoch is set to $3$ and we choose the 

The hyper-parameters for collecting step-level preference data via MCTS and training Q-value models are summarized in Table~\ref{appendix:tab:hyper-parameters}.

\begin{table}[h]%
	\centering
	\caption{Hyper-parameters for our experimental results. We used nearly identical hyper-parameters for both tasks. Where differences exist, the value for the WebShop task are listed first, followed by that for HotPotQA.}
	\vskip 0.1in
	%	\resizebox{\linewidth}{!}{
		% \setlength{\tabcolsep}{1.5mm}{
			\begin{tabular}{l|cc}
				\toprule
				\textbf{Stage}& \textbf{Hyper-parameter}  & \textbf{Value} \\
				\cmidrule(lr){1-3} % 调整中间横线长度
				\multirow{6}{*}{MCTS}&maximum iterations $m$& 30 \\
				&sampling number $n$ & 5  \\
				%				&\begin{tabular}[c]{@{}c@{}}max length for \\[-0.5ex] LLM-as-a-judge instructions\end{tabular}  & 768 & 1124\\
				& exploration weight $\eta$ & 2  \\
				&sampling temperature& 1  \\
				&maximum depth & 10, 7 \\
				\cmidrule(lr){1-3}
%				\multirow{6}{*}{\begin{tabular}[l]{@{}l@{}}Training Q-value \\[-0.5ex] Models\end{tabular}}& KL weighting parameter $\beta$ &  0.1  \\
				\multirow{5}{*}{Training Q-value Models}& KL weighting parameter $\beta$ &  0.1  \\
				& warm-up ratio & 0.1  \\
				&learning rate& 1e-5  \\
				&batch size& 16  \\
				&max context length & 4096  \\
				% &\begin{tabular}[c]{@{}c@{}}preference-strengthening \\[-0.5ex] coefficient $\lambda$ \end{tabular}  & 0.01  & & ??? \\
				\bottomrule
			\end{tabular}
			%		}
		\label{appendix:tab:hyper-parameters}
	\end{table}

\section{Case Study on WebShop}
In this section, we present a case study to further analyze the action selection guided by Q-value models.
We first show the ReAct-style instruction prompt in Figure~\ref{fig:webshop_prompt}.
Then we compare the decision-making process of the GPT-4o-mini agent with and without the Q-value model on WebShop in Figure~\ref{fig:webshop_case}.

\label{appendix:webshop-case}
\begin{tcolorbox}[breakable,title=Instruction Prompt for WebShop ]
	You are doing a web shopping task.
	I will give you instructions about what to do.
	You have to follow the instructions. Every round I will give you an observation and a list of available actions, you have to respond to an action based on the state and instruction.
	You can use search action if search is available.
	You can click one of the buttons in clickables.
	An action should be one of the following structure: search[keywords] or click[value]\\
	
	If the action is not valid, perform nothing.
	Keywords in search are up to you, but the value in click must be a value in the list of available actions.
	Remember that your keywords in search should be carefully designed.\\
	
	Your response should use the following format:\\
	Thought: I think ...\\
	Action: click[something]
\end{tcolorbox}
\begin{figure}[ht]
	\centering
	\vspace{-8pt}
	\caption{
		Instruction prompt for WebShop.
	}
	\label{fig:webshop_prompt}
\end{figure}

\newpage

\begin{tcolorbox}[breakable,title=Case Study on WebShop]
	\columnseprule=0.5pt
	\begin{multicols}{2}
		\begin{center}\textcolor{red}{\textbf{\Large Greedy Decision-making (part 1)}}\end{center}
%		\textcolor{dm-blue-500}{\textit{STEP 0}} \\
		\textcolor{case_purple}{\textbf{Instruction:}} WebShop [SEP] Instruction: [SEP] i am looking for anti slip women sandals. please choose black one, and price lower than 50.00 dollars [SEP] Search \\
		\textcolor{case_blue}{\textbf{Thought:}}I think I should search for black anti-slip women sandals within the specified price range. \\
		\textcolor{case_blue}{\textbf{Action:}} search[black anti slip women sandals price $<$ 50.00]\\
%		\textcolor{dm-blue-500}{\textit{STEP 1}} \\
		\textcolor{case_purple}{\textbf{Observation:}}  \textit{$<$searching results$>$}\\
		\textcolor{case_blue}{\textbf{Thought:}}  \sethlcolor{red!50}\hl{I think I should click on the product that fits the criteria of being black and under \$50.00.} \\
		\textcolor{case_blue}{\textbf{Action:}} \sethlcolor{red!50}\hl{click[B09BJB5JR5]} \\
		\textcolor{case_purple}{\textbf{Observation:}} Instruction: [SEP] i am looking for anti slip women sandals. please choose black one, and price lower than 50.00 dollars [SEP] Back to Search [SEP] $<$ Prev [SEP] Shoes for Women Casual Women's Classic Comfy Pointed Toe Pearl Flat Pumps Shoes Slip-on Travel Shoes Black [SEP] Price: \$100.0 [SEP] Rating: N.A. [SEP] Description [SEP] Features [SEP] Reviews [SEP] Buy Now" \\
		\textcolor{case_blue}{\textbf{Thought:}} I realize that the selected item is not an anti-slip sandal and is over the price limit. I need to look for a different option that fits the criteria.  \\
		\textcolor{case_blue}{\textbf{Action:}} nothing \\
		\textcolor{case_purple}{\textbf{Observation:}} Instruction: [SEP] i am looking for anti slip women sandals. please choose black one, and price lower than 50.00 dollars [SEP] Back to Search [SEP] $<$ Prev [SEP] Shoes for Women Casual Women's Classic Comfy Pointed Toe Pearl Flat Pumps Shoes Slip-on Travel Shoes Black [SEP] Price: \$100.0 [SEP] Rating: N.A. [SEP] Description [SEP] Features [SEP] Reviews [SEP] Buy Now \\
		\textcolor{case_blue}{\textbf{Thought:}} Since the previous options did not meet the criteria for anti-slip black women sandals under \$50.00, I will need to search again to find suitable options. \\
		\textcolor{case_blue}{\textbf{Action:}} perform nothing. \\
		\textcolor{case_purple}{\textbf{Observation:}} Instruction: [SEP] i am looking for anti slip women sandals. please choose black one, and price lower than 50.00 dollars [SEP] Back to Search [SEP] $<$ Prev [SEP] Shoes for Women Casual Women's Classic Comfy Pointed Toe Pearl Flat Pumps Shoes Slip-on Travel Shoes Black [SEP] Price: \$100.0 [SEP] Rating: N.A. [SEP] Description [SEP] Features [SEP] Reviews [SEP] Buy Now \\
		\textcolor{case_blue}{\textbf{Thought:}} As I still do not see any options for anti-slip black sandals under \$50.00, I will attempt to search again to find the right items.  \\
		\columnbreak
		
		\begin{center}\textcolor{my_green}{\textbf{\Large Guiding Action Selection with Q  (part 1)}}\end{center}
		\textcolor{case_purple}{\textbf{Instruction:}} WebShop [SEP] Instruction: [SEP] i am looking for anti slip women sandals. please choose black one, and price lower than 50.00 dollars [SEP] Search \\
		\textcolor{case_blue}{\textbf{Thought:}} I think I should search for black anti-slip women sandals within the specified price range. \\
		\textcolor{case_blue}{\textbf{Action:}} search[black anti-slip women sandals price $<$ 50.00]\\
		\textcolor{case_purple}{\textbf{Observation:}}  \textit{$<$searching results$>$}\\
		\textcolor{case_blue}{\textbf{Thought 1:}} \sethlcolor{green!50}\hl{I think I should select the Gibobby Platform Sandals for Women, which are anti-slip and fit the price criteria.} \\
		\textcolor{case_blue}{\textbf{Action 1:}} \sethlcolor{green!50}\hl{click[B07TN8VHKC] $ (Q = -8.8607)$ }\\
		\textcolor{case_blue}{\textbf{Thought 2:}} \sethlcolor{gray!50}\hl{I think I should select the platform flip flop option that is under 50.00 dollars, as it seems to fit my requirements for black anti-slip sandals.} \\
		\textcolor{case_blue}{\textbf{Action 2:}}  \sethlcolor{gray!50}\hl{click[B09S8BF2GR] $ (Q = -9.2938)$} \\
		\textcolor{case_blue}{\textbf{Thought 3:}} \sethlcolor{gray!50}\hl{I think I should click on the B09BJB5JR5 product because it is a black shoe option, but it exceeds my price criteria. None of the sandals listed are black anti-slip options under \$50. I'll look for the most relevant product within budget and parameters.}\\
		\textcolor{case_blue}{\textbf{Action 3:}} \sethlcolor{gray!50}\hl{ click[B09S8BF2GR]  $ (Q = -9.1830)$} \\
		\textcolor{case_blue}{\textbf{Thought 4:}} \sethlcolor{gray!50}\hl{ I see there are options for sandals under 50.00 dollars; the Gibobby Slippers for Women is an anti-slip option. I should click on that. }\\
		\textcolor{case_blue}{\textbf{Action 4:}} \sethlcolor{gray!50}\hl{ click[B07TN7SRFN]  $ (Q = -9.0580)$} \\
		\textcolor{case_blue}{\textbf{Thought 5:}} \sethlcolor{gray!50}\hl{ I think I should select one of the black sandals that fits the criteria and is priced under 50.00 dollars. }\\
		\textcolor{case_blue}{\textbf{Action 5:}} \sethlcolor{gray!50}\hl{  click[B09S8BF2GR]  $ (Q = -9.6835)$ }\\
		\textcolor{case_purple}{\textbf{Observation:}} Instruction: [SEP] i am looking for anti slip women sandals. please choose black one, and price lower than 50.00 dollars [SEP] Back to Search [SEP] color [SEP] a1-black [SEP] beige [SEP] pink [SEP] yellow [SEP] z2-red [SEP] \textit{$<$More product details...$>$}   Gibobby Platform Sandals for Women Summer,Womens Bows Flip Flops Flat Sandals Anti-Slip Casual Beach Thong Slipper [SEP] Price: $0.01 to $16.71 [SEP] Rating: N.A. [SEP] Description [SEP] Features [SEP] Reviews [SEP] Buy Now\\
		\textcolor{case_blue}{\textbf{Thought:}} \sethlcolor{green!50}\hl{I think I should select the black color option for the Gibobby Platform Sandals.}\\
		\textcolor{case_blue}{\textbf{Action:}} \sethlcolor{green!50}\hl{click[a1-black]}\\
	\end{multicols}
\end{tcolorbox}
\newpage
\begin{tcolorbox}[breakable,title=Case Study on WebShop]
	\columnseprule=0.5pt
	\begin{multicols}{2}
		\begin{center}\textcolor{red}{\textbf{\Large Greedy Decision-making (part 2)}}\end{center}
		\textcolor{case_blue}{\textbf{Action:}} perform nothing. \\
		\textcolor{case_purple}{\textbf{Observation:}} Instruction: [SEP] i am looking for anti slip women sandals. please choose black one, and price lower than 50.00 dollars [SEP] Back to Search [SEP] < Prev [SEP] Shoes for Women Casual Women's Classic Comfy Pointed Toe Pearl Flat Pumps Shoes Slip-on Travel Shoes Black [SEP] Price: \$100.0 [SEP] Rating: N.A. [SEP] Description [SEP] Features [SEP] Reviews [SEP] Buy Now \\
		\textcolor{case_blue}{\textbf{Thought:}} It appears that there are no suitable options available on the current page for black anti-slip women sandals under \$50.00.  \\
		\textcolor{case_blue}{\textbf{Action:}} perform nothing. \\
		\\
		\textit{$<$Several repeated actions......$>$} \\
		\\
		\textcolor{case_purple}{\textbf{Reward:}} 0 \\
		\\
		\sethlcolor{red!50}\hl{\textit{\#  Prioritizing ``black" and ``price" does not meet all criteria, leading to task failure.}}
		\\
		\\
		\\
		\\
		\\
		\\
		\columnbreak
		
		\begin{center}\textcolor{my_green}{\textbf{\Large Guiding Action Selection with Q (part 2)}}\end{center}
		\textcolor{case_purple}{\textbf{Observation:}} Instruction: [SEP] i am looking for anti slip women sandals. please choose black one, and price lower than 50.00 dollars [SEP] Back to Search [SEP] $<$ Prev [SEP] color [SEP] a1-black [SEP] beige [SEP] pink [SEP] yellow [SEP] z2-red [SEP]\textit{$<$More product details...$>$} Gibobby Platform Sandals for Women Summer,Womens Bows Flip Flops Flat Sandals Anti-Slip Casual Beach Thong Slipper [SEP] Price: $0.01 to $16.71 [SEP] Rating: N.A. [SEP] Description [SEP] Features [SEP] Reviews [SEP] Buy Now \\
		\textcolor{case_blue}{\textbf{Thought:}} I think I should click on the 'Buy Now' button to purchase the black Gibobby Platform Sandals.\\
		\textcolor{case_blue}{\textbf{Action:}} click[Buy Now]\\
		\textcolor{case_purple}{\textbf{Observation:}} Thank you for shopping with us! [SEP] Your code: [SEP] None [SEP] (Paste it in your MTurk interface.) [SEP] Purchased [SEP] asin [SEP] B07TN8VHKC [SEP] options [SEP] \{``color": ``a1-black"\} [SEP] attrs [SEP] \textit{$<$More purchasing details...$>$} Reward [SEP] Your score (min 0.0, max 1.0) [SEP] 1.0 [SEP] Reward Details [SEP] None\\ 
		\\
		\textcolor{case_purple}{\textbf{Reward:}} 1.0 \\
		\\
		\sethlcolor{green!50}\hl{\textit{\# Prioritizing other attributes and then selecting ``black" satisfies all criteria.}}
		\\
	\end{multicols}
\end{tcolorbox}
\begin{figure}[ht]
	\centering
	\vspace{-8pt}
	\caption{
		Case study on WebShop. 
		To analyze the action selection guided by our Q-value model, we take the second step of the decision-making process as an example. We present multiple sampled candidate actions along with their corresponding Q-values. Action 1, which prioritizes the attributes ``women" ``anti-slip" and ``price", has the highest Q-value among the five candidates and is therefore selected in the second decision-making step. The ``black" attribute is satisfied at the next step.
	}
	\label{fig:webshop_case}
\end{figure}

\section{Case Study on HotPotQA}
\label{appendix:case:HotPotQA}

\begin{tcolorbox}[breakable,title=Instruction Prompt for HotPotQA ]
	Solve a question answering task with interleaving Thought, Action, Observation steps. Thought can reason about the current situation, and Action can be three types: \\
	\\
	(1) Search[entity], which searches the exact entity on Wikipedia and returns the first paragraph if it exists. If not, it will return some similar entities to search.\\
	(2) Lookup[keyword], which returns the next sentence containing keyword in the current passage.\\
	(3) Finish[answer], which returns the answer and finishes the task.\\
	\\
	After each observation, provide the next Thought and next Action.
\end{tcolorbox}
\begin{figure}[ht]
	\centering
	\vspace{-8pt}
	\caption{
		Instruction prompt for HotPotQA.
	}
	\label{fig:hotpotQA_prompt}
\end{figure}

\newpage

\begin{tcolorbox}[breakable,title=Case Study on HotPotQA]
	\columnseprule=0.5pt
	\begin{multicols}{2}
		\begin{center}\textcolor{red}{\textbf{\Large Greedy Decision-making (part 1)}}\end{center}
		\textcolor{case_purple}{\textbf{Instruction:}} Question: The season in which ``The Death Camp of Tolerance" was an episode in began airing on what date? \\
		\textcolor{case_blue}{\textbf{Thought:}} I need to find information about the episode ``The Death Camp of Tolerance" to determine which season it is in and when that season began airing.\\
		\textcolor{case_blue}{\textbf{Action:}} Search[``The Death Camp of Tolerance" episode air date] \\
		\textcolor{case_purple}{\textbf{Observation:}}  Could not find ``The Death Camp of Tolerance" episode air date. Similar: ['The Death Camp of Tolerance', 'South Park season 6 (category South Park season 6 episodes)', 'List of South Park home video releases', 'Awesom-O (category Television episode articles with short description for single episodes)', 'Bass to Mouth (category Television episode articles with short description for single episodes)'].\\
		\textcolor{case_blue}{\textbf{Thought:}} The search didn't return the specific air date for ``The Death Camp of Tolerance." I know it belongs to ``South Park," so I should search for the sixth season of South Park and find its premiere date.\\
		\textcolor{case_blue}{\textbf{Action:}}  Search[``South Park season 6 air date"]\\
		\textcolor{case_purple}{\textbf{Observation:}} There were no results matching the query.. The pages ``South Park season 6 air date" and ``South Park season 6 air date" do not exist. You can create a draft and submit it for review or request that a redirect be created. \\
		\textcolor{case_blue}{\textbf{Thought:}} I couldn't find specific information about the sixth season of South Park, but I know ``The Death Camp of Tolerance" is from Season 6, which aired in the early 2000s. I will look up the general air date for South Park Season 6.\\
		\textcolor{case_blue}{\textbf{Action:}}  Search[South Park Season 6 premiere date]\\
		\textcolor{case_purple}{\textbf{Observation:}}  Could not find South Park Season 6 premiere date. Similar: ['South Park season 26', 'Cobra Kai season 6', 'List of South Park episodes', 'South Park season 20', 'South Park season 19'].\\
		\textcolor{case_blue}{\textbf{Thought:}} Since I am unable to find information through direct searches, I will try to look for any details surrounding general information about ``The Death Camp of Tolerance" episode itself as it may include air dates or related context.\\
		\textcolor{case_blue}{\textbf{Action:}}  Search[``The Death Camp of Tolerance South Park episode details"] \\
		\textcolor{case_purple}{\textbf{Observation:}}  There were no results matching the query.. The pages ``The Death Camp of Tolerance South Park episode details" and ``The Death Camp of Tolerance South Park episode details" do not exist. You can create a draft and submit it for review or request that a redirect be created.\\
		\textcolor{case_blue}{\textbf{Thought:}} Since I am unable to find relevant pages regarding the episode or the season via search, I can infer    \\
		\columnbreak
		
		\begin{center}\textcolor{my_green}{\textbf{\Large Guiding Action Selection with Q (part 1)}}\end{center}
		\textcolor{case_purple}{\textbf{Instruction:}} Question: The season in which ``The Death Camp of Tolerance" was an episode in began airing on what date? \\
		\textcolor{case_blue}{\textbf{Thought:}} I need to search for ``The Death Camp of Tolerance" to find out which season it belongs to and when that season began airing. \\
		\textcolor{case_blue}{\textbf{Action:}} Search[The Death Camp of Tolerance] \\
		\textcolor{case_purple}{\textbf{Observation:}}  ``The Death Camp of Tolerance" is the fourteenth episode of the sixth season of the American animated television series South Park, and the 93rd overall episode of the series. It originally aired on Comedy Central in the United States on November 20, 2002.. In the episode, Mr. Garrison tries to get fired from his new job as the fourth grade teacher at South Park Elementary by being overtly and explicitly homosexual in front of the students, so he can sue the school and get twenty-five million dollars in damages. However, he is unable to as the rest of the faculty and the children's parents are desperate to appear tolerant of his homosexuality.\\
		\textcolor{case_blue}{\textbf{Thought:}} ``The Death Camp of Tolerance" is the fourteenth episode of the sixth season and aired on November 20, 2002. I need to find out when the sixth season began airing. \\
		\textcolor{case_blue}{\textbf{Action:}} Search[sixth season of South Park] \\
		\textcolor{case_purple}{\textbf{Observation:}}  Could not find sixth season of South Park. Similar: ['South Park season 26', 'South Park season 6', ``Bebe's Boobs Destroy Society (category South Park season 6 episodes)", 'South Park season 7', 'List of South Park episodes'].\\
		\textcolor{case_blue}{\textbf{Thought:}}  I couldn't find specific information about when the sixth season of South Park began airing directly. However, I can try looking up just ``South Park season 6" to get information on the entire season.  \\
		\textcolor{case_blue}{\textbf{Action:}}  Search[South Park season 6] \\
		\textcolor{case_purple}{\textbf{Observation:}}  The sixth season of South Park, an American animated television series created by Trey Parker and Matt Stone, began airing on March 6, 2002. The sixth season concluded after 17 episodes on December 11, 2002.[1]. The sixth season is the only season of South Park to not feature Kenny McCormick as a main character, due to him being killed off at the end of the previous season in ``Kenny Dies." He was revived at the end of the season finale, ``Red Sleigh Down", appearing in the last few seconds. This was his sole appearance in the season.. Kenny was written off in ``Kenny Dies" at the end of season 5 of South Park, though he was mentioned throughout season 6. \\
				\textcolor{case_blue}{\textbf{Thought:}} The sixth season of South Park began airing on March 6, 2002. ``The Death Camp of Tolerance" aired on November 20, 2002, during that season. Therefore, the answer to when the season began airing is March 6, 2002.\\
	\end{multicols}
\end{tcolorbox}
\begin{tcolorbox}[breakable,title=Case Study on HotPotQA]
	\columnseprule=0.5pt
	\begin{multicols}{2}
		\begin{center}\textcolor{red}{\textbf{\Large Greedy Decision-making (part 2)}}\end{center}
		that ``The Death Camp of Tolerance" is recognized more broadly and may have been aired sometime in 2002. I need to make an educated guess based on the known timeframe of ``South Park." \\
		\textcolor{case_blue}{\textbf{Action:}} Finish[2002] \\
		\\
		\textcolor{case_purple}{\textbf{Reward:}} 0 \\
		\\
		\sethlcolor{red!50}\hl{\textit{\#  The generated keywords are insufficiently accurate for searching.}}
		\columnbreak
		
		\begin{center}\textcolor{my_green}{\textbf{\Large Guiding Action Selection with Q (part 2)}}\end{center}
		\textcolor{case_blue}{\textbf{Action:}}  Finish[March 6, 2002]\\
		\\
		\textcolor{case_purple}{\textbf{Reward:}} 1 \\ 
		\\
		\sethlcolor{green!50}\hl{\textit{\# Guided by the Q-value model, the selected search keywords are better suited for this task.}}
	\end{multicols}
\end{tcolorbox}
\begin{figure}[ht]
	\centering
	\vspace{-8pt}
	\caption{
		Case study on HotPotQA. 
	}
	\vspace{-8pt}
	\label{fig:hotpotqa_case}
\end{figure}

\section{Pseudocode}
Our method comprises both training and inference stages. During the training stage, we first use Monte Carlo Tree Search (MCTS) to explore high-quality trajectories, annotating the actions in each step with Q-values. We then construct preference data and train the Q-value model using step-level Direct Policy Optimization (DPO). During inference, the trained Q-value model guides action selection at each decision-making step.

\begin{algorithm}
	\caption{Training of Q-value models.}
	\label{alg:training}
	\begin{algorithmic}
		\STATE \textbf{Input:} $\pi$: LLM agent policy; $\pi_\mathrm{ref}$: initial LLM for training the Q-value model; $m$: number of iterations; $n$: number of candidate actions to sample; $T$: maximum MCTS tree depth and environment step.
		\STATE $\pi_{\theta} \leftarrow \pi_{\text{ref}}$
%		\STATE Sample a batch of $B$ tasks from $\mathcal{D}_T$
%		\STATE \COMMENT{Stage 1: Training}
		\FOR{each task instruction in the training set}  
		\STATE Initialize the root node $s_0$
		\FOR{each MCTS iteration} 
		\FOR{$t \gets 0, \dots, T$}
		\IF{$s_t$ is not terminal}
		% wip
%		\FOR{$i \gets 0, \dots, n$}
%		\STATE Sample $a_t^{(i)} \sim \pi(s_t)$
%		\ENDFOR
		\STATE \textbf{Expansion}: Sample $n$ candidate actions $a_t \sim \pi(s_t)$, and obtain corresponding observation $o_t$ from the environment.
		\STATE \textbf{Evaluation}: From the selected node's trace, roll out the trajectory using $\pi$ until a terminal state is reached
		\STATE \textbf{Backpropagation:} Backpropagate the value estimate bottom-up according to Equation~\ref{eq:back}.
		\STATE \textbf{Selection:} Traverse tree from the root node to a leaf node according to UCT in Equation~\ref{eq:uct}.
		\ENDIF
		\ENDFOR
		\ENDFOR
%		\STATE Collect trajectories from rollouts and store them in replay buffer $\mathcal{B}$
		\ENDFOR
		\STATE Construct preference pairs $\mathcal{D} = \{u, \tau_t, a_t^w, a_t^l\}_{t=1}^{T}$ according to the final trees.
		\STATE Optimize $\pi_{\theta}$ using step-level DPO objective in Equation~\ref{eq:step-dpo} with $\mathcal{D}$.
		\STATE \textbf{Output:} $\pi_{\theta}$, the well-trained Q-value models
	\end{algorithmic}
\end{algorithm}

	\vspace{-16pt}
\begin{algorithm}
	\caption{Inference with Q-value models.}
	\label{alg:inference}
	\begin{algorithmic}
		\STATE \textbf{Input:} $\pi_\mathrm{ref}$: initial LLM for training the Q-value model; $\pi_{\theta}$: well-trained Q-value models; $n$: number of candidate actions to sample; $T$: maximum MCTS tree depth and environment step.
		%		\STATE Sample a batch of $B$ tasks from $\mathcal{D}_T$
		%		\STATE \COMMENT{Stage 1: Training}
		\FOR{each task instruction in the test set}  
			\FOR{$t \gets 0, \dots, T$}
				\IF{$s_t$ is not terminal}
					\STATE Sample $n$ candidate actions $a \sim \pi(s_t)$, and calculate the $Q(u, \tau_t, a_t)$ according to Equation~\ref{eq:Q-value}.
					\STATE Select the action $a_t = \arg\max_{a}\Bigl[Q(u, \tau_t, a)\Bigl]$ to interact with the environment.
				\ENDIF
			\ENDFOR
		\ENDFOR
	\end{algorithmic}
\end{algorithm}